\documentclass[10pt,journal,compsoc]{IEEEtran}
\ifCLASSOPTIONcompsoc
  \usepackage[nocompress]{cite}
\else
  \usepackage{cite}
\fi
\ifCLASSINFOpdf
\else
\fi

\usepackage{url}
\usepackage{amssymb}
\usepackage{amsmath}
\usepackage{algorithm}
\usepackage{algorithmicx}
\usepackage{algpseudocode}
\usepackage{mathrsfs}
\usepackage{bbm}
\usepackage{booktabs}
\usepackage{tabularx}
\usepackage{multirow}
\usepackage{makecell}
\usepackage{multicol}
\usepackage{graphicx}
\usepackage{subfig}
\usepackage{hyperref}
\usepackage{pifont}
\usepackage{bbding}
\usepackage{rotating}
\newtheorem{remark}{Remark}
\usepackage{color}
\usepackage{colortbl}
\usepackage{diagbox}
\definecolor{Gray}{gray}{0.9}
\definecolor{LGray}{gray}{0.8}

\graphicspath{{images/}}

\begin{document}
%
\title{Reliable Representation Learning for Incomplete Multi-View Missing Multi-Label Classification}
%
%
%
%

\author{Chengliang Liu,
        Jie Wen*, \emph{Senior Member}, \emph{IEEE},
        Yong Xu*, \emph{Senior Member}, \emph{IEEE},\\
        Bob Zhang, \emph{Senior Member}, \emph{IEEE},
        Liqiang Nie, \emph{Senior Member}, \emph{IEEE},
        Min Zhang
\IEEEcompsocitemizethanks{\IEEEcompsocthanksitem This work is partially supported by the Shenzhen Science and Technology Program under Grant No. JCYJ20240813105135047, National Natural Science Foundation of China under Grant No. 62372136, and Guangdong Basic and Applied Basic Research Foundation under Grant No. 2024A1515030213.
\IEEEcompsocthanksitem Chengliang Liu and Jie Wen are with the Shenzhen Key Laboratory of Visual Object Detection and Recognition, Harbin Institute of Technology, Shenzhen 518055, China. Email: liucl1996@163.com; jiewen\_pr@126.com.\protect\\
\IEEEcompsocthanksitem Yong Xu is with the Shenzhen Key Laboratory of Visual Object Detection and Recognition, Harbin Institute of Technology, Shenzhen 518055, China and the Pengcheng Laboratory, Shenzhen 518055, China. Email: laterfall@hit.edu.cn.\protect\\
\IEEEcompsocthanksitem Bob Zhang is with the PAMI Research Group, Department of Computer and Information Science, University of Macau, Macau, China. Email: bobzhang@um.edu.mo.
\IEEEcompsocthanksitem Liqiang Nie and Min Zhang are with the School of Computer Science and Technology, HarbinInstitute of Technology, Shenzhen 518055, China. Email: nieliqiang@gmail.com; zhangmin2021@hit.edu.cn.
\IEEEcompsocthanksitem Corresponding authors: Jie Wen and Yong Xu. Email: jiewen\_pr@126.com; laterfall@hit.edu.cn.
}
}

%
%

\markboth{Journal of \LaTeX\ Class Files,~Vol.~14, No.~8, August~2015}%
{Shell \MakeLowercase{\textit{et al.}}: Bare Demo of IEEEtran.cls for Computer Society Journals}
%



\IEEEtitleabstractindextext{%
\begin{abstract}
As a cross-topic of multi-view learning and multi-label classification, multi-view multi-label classification has gradually gained traction in recent years. The application of multi-view contrastive learning has further facilitated this process, however, the existing multi-view contrastive learning methods crudely separate the so-called negative pair, which largely results in the separation of samples belonging to the same category or similar ones. Besides, plenty of multi-view multi-label learning methods ignore the possible absence of views and labels. To address these issues, in this paper, we propose an incomplete multi-view missing multi-label classification network named RANK. In this network, a label-driven multi-view contrastive learning strategy is proposed to leverage supervised information to preserve the intra-view structure and perform the cross-view consistency alignment. Furthermore, we break through the view-level weights inherent in existing methods and propose a quality-aware sub-network to dynamically assign quality scores to each view of each sample. The label correlation information is fully utilized in the final multi-label cross-entropy classification loss, effectively improving the discriminative power. Last but not least, our model is not only able to handle complete multi-view multi-label data, but also works on datasets with missing instances and labels. Extensive experiments confirm that our RANK outperforms existing state-of-the-art methods.
\end{abstract}

}

\maketitle

\IEEEdisplaynontitleabstractindextext

%
\IEEEpeerreviewmaketitle

\IEEEraisesectionheading{\section{Introduction}\label{sec:introduction}}

%
%
%
%
\IEEEPARstart{A}{s} a classic classification task, multi-label classification better matches the complex and diverse application requirements in the real environment than single-label classification \cite{zhang2013review,yu2021multi}. For example, we are used to labeling a landscape picture with multiple elements -- "sky", "mountain", "flying bird"; or describing the attributes of a piece of news with a specific set of words: "race", "football" ,"sports" \cite{li2022learning,patel2022modeling}. However, unlike single-label classification tasks that require mutual exclusivity among labels, multi-label data naturally maintains complex label correlations, which enriches the semantic information of multi-label data but also increases the difficulty of multi-label learning.

Going to our topic, multi-view multi-label classification (MvMLC) is driven by an emerging data form: multi-view data. Thanks to the development of various information collection technologies, we no longer describe the target from a single perspective, that is, multiple heterogeneous features jointly represent the observation subjects \cite{wen2022survey,wang2023flora,fang2024your}. As a complex composite task, MvMLC maintains the characteristics of multi-view learning and multi-label classification at the same time, which need to be considered comprehensively in a unified framework \cite{liu2023dicnet}.

From the perspective of multi-view representation learning, on the one hand, multi-view consistency is viewed as an important modeling criterion, i.e., we expect to obtain a unique representation learning result from multiple views \cite{fang2021animc}. To accomplish this goal, in recent years, some advanced methods adopt contrastive learning to aggregate multiple views, the purpose of which is to maximize the similarity or mutual information of instances in different views to satisfy consistency \cite{xu2022multi,pan2021multi,lin2021completer}. However, such a design inevitably leads to category conflicts no matter in clustering or classification tasks, because it is difficult to guarantee that the negative pairs selected are truly negative pairs, and blindly pulling apart these so-called negative pairs is likely to increase the intra-class distance \cite{huang2022learning}. To solve this problem and extract reliable embedding feature, in this paper, we design a label-driven multi-view contrastive learning strategy that abandons pushing away negative pairs and instead guides instances to hold an appropriate relative distance, so that it can maintain cross-view consistency while avoid breaking inter-sample relationships. On the other hand, multi-view complementarity is regarded as the basis for the success of multi-view learning. Some studies implicitly consider this complementarity in cross-view information fusion \cite{trosten2021reconsidering,kang2020multi}, meanwhile, numerous works attempt to introduce the same balance factor for different views of all samples, roughly defining which view is better \cite{li2015large,liu2022localized}. However, we argue that the discrimination of different samples in the same view is different. Considering such a sample-level view discrepancy, in this paper, we propose a quality-aware sub-network to explicitly exploit multi-view complementarity to obtain the reliable sample-level fusion representation.

From the perspective of multi-label classification, one category often co-occurs with other categories, such as "mountain"-"sky" and "bird"-"tree" usually co-occur in the same image, which is called label correlations (or label dependencies) and this inter-category correlation is unsymmetrical \cite{ma2021label,chen2019multi}. Different from single-label classification, multi-label correlation plays an essential role in promoting efficient classification and has been paid increasing attention in lots of multi-label classification methods. In our paper, we develop a novel multi-label collaborative cross-entropy loss to replace the traditional multi-label binary cross-entropy loss \cite{ma2021label,hang2021collaborative}, by introducing the global label correlation information, which breaks the inter-class barrier existing in conventional losses.

In addition, a growing number of researchers believe that the completeness assumption of multi-view multi-label data is hard to adapt to real scenarios. Data-missing is unavoidable no matter with heterogeneous features or multi-label tags, and this incompleteness has a huge blow to the MvMLC methods designed for ideal scenario. In view of this, we extend our method to incomplete multi-view missing multi-label classification (iM3C) task. Unlike most existing incomplete MvMLC methods that are unable to handle both missing views and labels, our method is allowed to flexibly cope with arbitrary missing cases, greatly enhancing its robustness. Meanwhile, we perform extensive experiments to verify the effectiveness and advanced nature of our method under both complete and incomplete conditions.

In summary, we named our method \textbf{R}eliable represent\textbf{A}tion lear\textbf{N}ing networ\textbf{K} (\textbf{RANK} for short) and distill our contributions as follows:
\begin{itemize}
\item We propose a label-driven multi-view contrastive learning strategy that preserves multi-view consensus alignment while abandoning traditional negative pair separation. The multi-label supervision information is fully mined to preserve the geometric structure within the view.
\item We propose a sample-level view quality-aware sub-network, which effectively helps the classification network to learn reliable cross-view fusion representations by exploiting multi-view complementarity explicitly.
\item We propose a novel multi-label collaborative cross-entropy loss, which utilizes global category dependencies to compute cross-class cross-entropy, rather than computing multiple independent binary cross-entropies as existing deep methods do.
\item Our RANK adopts a simple but effective approach to accommodate multi-view multi-label data with random missing. Extensive experiments performed in both complete and incomplete cases confirm the advances and robustness of our method.
\end{itemize}
 

\section{Related works}
\label{sec2}
Since our target task is closely related to the two domains, multi-label classification and multi-view learning, in this section, we sequentially introduce related works in multi-label classification, multi-view learning, MvMLC, and iM3C.
\subsection{Multi-label Classification}
For a long time, multi-label classification has received continuous attentions from plenty of scholars in machine learning. In a nutshell, the researches on multi-label classification can be roughly divided into three strategies \cite{zhang2013review}: 1) The multi-label classification task is simply regarded as multiple independent binary classification problems, ignoring the correlation among labels, such as ML-kNN \cite{zhang2007ml} and LIFT \cite{zhang2014lift}. 2) The correlation of two paired labels is utilized, considering the ordering relationship of labels or introducing a correlation matrix constructed by priori knowledge, such as CRPC \cite{furnkranz2008multilabel} and DMLkNN \cite{younes2011dependent}. 3) Mining the global label correlation, that is, considering that each label interacts with other labels \cite{cheng2009combining,yan2007model}. Apparently, the third strategy is more conducive to modeling complex label correlation but has higher complexity \cite{zhang2013review}. In general, most multi-label classification works are built on exploiting second-order or high-order label correlations. Huang et al. proposed LLSF, which associates each category to a feature subset and converts label correlations to feature subset correlations \cite{huang2015learning}. Chen et al. developed a graph convolutional neural network for multi-label image classification, which maps the label text to the word embedding space, and incorporates the pre-constructed label correlation matrix into the label-specific representation learning \cite{chen2019multi}.
\subsection{Multi-view Learning}
Multi-view learning can be combined with numerous downstream tasks. According to the presence or absence of supervised information, multi-view learning can be simply divided into supervised multi-view learning (e.g., multi-view classification) and unsupervised multi-view learning (e.g., multi-view clustering). Almost all multi-view learning methods revolve around the two basic assumptions: multi-view consistency and complementarity. Nie et al. proposed a multi-view clustering framework with adaptive neighbor assignment, named MLAN, dynamically learning cross-view similarity matrix and performing spectral clustering for optimal results. This framework can be extended to semi-supervised classification tasks \cite{nie2017multi}. Li et al. jointed spectral embedding and low-rank tensor learning into a unified multi-view clustering framework, which synchronously learns the spectral embedding matrix and tensor representation \cite{li2021consensus}. Different from MLAN, this consensus graph learning method attempts to learn a common graph in the spectral embedding space instead of original feature space. Another interesting multi-view classification method is the TMC model proposed by Han et al., which calculates the uncertainty of samples on each view, helping fuses the cross-view classification results \cite{han2022trusted}. There is no doubt that this confidence-based fusion strategy can be seen as an explicit utilization of multi-view complementarity and endows the model with better reliability. In addition, Xu et al. proposed a representative deep multi-view clustering method (MFLVC), which employs the classic contrastive learning strategy to aggregate the deep features extracted from different views of the same sample, and push away different samples at the instance level \cite{xu2022multi}. Although this strategy largely guarantees cross-view consistency, it inevitably leads to category conflicts (i.e., samples belonging to the same cluster are indiscriminately separated in the feature space).
\subsection{MvMLC}
The MvMLC task combining multi-view learning and multi-label classification undoubtedly increases the complexity of the issue, since the corresponding methods need to take into account the characteristics of each field at the same time. For instance, lrMMC is a typical matrix completion-based multi-view multi-label image classification model, which first learns low-rank common representations of all samples from multi-view features, and then utilizes the matrix completion method to obtain the test samples' labels \cite{liu2015low}. Another matrix factorization model called LSA-MML bridges the original features and the ground-truth label matrix through a common latent representation, and utilizes the Hilbert-Schmidt Independence Criterion (HSIC) \cite{gretton2005measuring} technique to align the basis matrices of different views in the kernel space to maximize the inter-view dependence\cite{zhang2018latent}. Focusing on both shared subspace mining and view-specific feature extraction, Wu et al. proposed a deep MvMLC network named SIMM. It forces view-specific features to be as different from common shared representation as possible to learn view-private information \cite{wu2019multi}. All designs of lrMMC, LSA-MML, and SIMM ignore the label correlation. Zhao et al. developed a simple neural network-based method to mine multi-view consistency and diversity (CDMM), which predicts each view's result and obtain the final prediction through late fusion \cite{zhao2021consistency}. Unlike LAS-MML, CDMM reversely utilizes HSIC to induce diversity across different views and a symmetric label dependency matrix is introduced. LVSL is a complex method for non-aligned multi-view multi-label classification task, which applies the Laplacian graph regularization constructed from the feature space in the process of learning view-specific labels to maintain the geometric structure of the raw data. Furthermore, a low-rank constraint is imposed on the learnable label correlation matrix considering the local label low-rank structure \cite{zhao2022non}.
\subsection{iM3C}

As the name suggests, the incomplete issue in MvMLC contains two sub-problems of missing views and labels. In recent years, incomplete multi-view learning have shown a good development momentum. For missing views, there are two main technical routes in existing methods, i.e., introducing the priori information to help the model ignore missing data or filling in missing data based on generation or interpolation approaches. For example, Liu et al. introduced a prior missing-view indicator matrix in the incomplete multi-view clustering task to help the model bypass missing instances \cite{liu2022localized}. SRLC proposed by Zhuge et al. constructs a similarity matrix based on incomplete data with missing location information, to measure the relationship between instance pairs \cite{zhuge2019simultaneous}. For the imputation route, Wen et al. attempted to align and infer missing views using graph embedding technique in a joint framework (UEAF) for efficient consensus representation learning \cite{wen2019unified}. Recformer developed by Liu et al seeks to recover missing views in an autoencoder framework and put the completed data into subsequent task flow \cite{liu2023information}. Similarly, in the multi-label classification domain, plenty of methods have been proposed to handle missing multi-label issue. GLOCAL, as a representative single-view multi-label classification model, applies both global and local label correlations to impose manifold regularization to the predicted labels respectively \cite{zhu2017multi}. Unlike other methods that introduce a priori label correlation matrix, GLOCAL directly learn the Laplace matrices corresponding to these label correlation matrices. In addition, a missing label indicator matrix is introduced to reduce the noise brought by missing labels \cite{zhu2017multi}. Another single-view multi-label classification model, DM2L, encourages the entire prediction matrix to be high-rank and the sub-label matrix to be low-rank to capture both global and local label structures. Similar to GLOCAL, DM2L also introduces an additional indicator matrix to help cope with incomplete labels \cite{ma2021expand}. Note that the lrMMC in last subsection is also compatible with missing labels due to its matrix completion technique. Although these studies have made outstanding contributions to incomplete multi-view learning or missing multi-label classification, they fail to accommodate both incomplete cases simultaneously. 

To the best of our knowledge, there are very few existing methods that can handle this double-incompleteness. For example, Tan et al. proposed an incomplete multi-view weak label learning model, iMVWL, which decomposes multi-view features into a latent common feature and multiple view-specific basis matrices, while connecting the latent feature space and label space through a mapping matrix. The two missing cases are considered separately by introducing the corresponding prior missing information \cite{tan2018incomplete}. After this, Li et al. deeply studied the global high-rank and local low-rank structures of multi-label matrix, and developed the non-aligned iM3C method termed NAIM3L, which avoids negative effects by constructing a composite index matrix involving available views and known label information \cite{li2022concise}. Recently, another work DICNet
proposed a novel deep iM3C framework with instance-level contrastive learning strategy, which aligns instances of samples on different views in an incomplete embedding space and pushes away instances belonging to different samples \cite{liu2023dicnet}. However, the multi-view contrastive learning adopted by DICNet still has the above-mentioned category conflict problem, so in this paper, we make a series of targeted improvements to it.
\begin{table}[!t]
\caption{Main notations used in this paper.}
\vspace{-0.2cm}
\centering
\small
\resizebox{\linewidth}{\height}{
	\begin{tabular}{|c|c|c|c|}
	    \toprule
	    $n$ & \makecell[c]{the number of \\samples} &$c$&\makecell[c]{the number of \\classes}\\ 
   		\midrule
   		$m$ & \makecell[c]{the number of \\views} &$d_v$&\makecell[c]{the dimension of \\ view $v$}\\ 
		\midrule
		$\mathbf{C}$ & \makecell[c]{label correlation\\ matrix ($[0,1]^{c\times c}$)} &$d_e$&\makecell[c]{the dimension of \\ embedding feature}\\ 
		\midrule
	     $\mathbf{X}^{(v)}$ & \makecell[c]{original data \\of view $v$ ($\mathbb{R}^{n\times d_v}$)} &$\mathbf{Y}$&\makecell[c]{label matrix\\ ($\{0,1\}^{n\times c}$)}\\ 
		\midrule
	    $\bar{\mathbf{X}}^{(v)}$ & \makecell[c]{reconstructed data\\ of view $v$ ($\mathbb{R}^{n\times d_v}$)}&$\mathbf{W}$&  \makecell[c]{missing-view \\ indicator ($\{0,1\}^{n\times m}$)}\\ 
	    \midrule
	    $\mathbf{Z}^{(v)}$ & \makecell[c]{embedding feature \\of view $v$ ($\mathbb{R}^{n\times d_e}$)}&$\mathbf{G}$&\makecell[c]{missing-label \\ indicator ($\{0,1\}^{n\times c}$)}\\ 
	    \midrule
	    $\bar{\mathbf{Z}}$ & \makecell[c]{fusion feature \\of all views ($\mathbb{R}^{n\times d_e}$)}&$\mathbf{L}$&\makecell[c]{similarity graph in\\ label space ($[0,1]^{n\times n}$)}\\ 
	    \midrule
 	    $\bar{\mathbf{F}}^{(v)}$ & \makecell[c]{similarity graph of\\ view $v$ in embeddi-\\ng space ($[0,1]^{n\times n}$)}&$\mathbf{B}$&\makecell[c]{predicted quality score\\ matrix by discrimina-\\tor ($[0,1]^{n\times m}$)}\\ 
 	    \midrule
 	    $\mathbf{Q}$ & \makecell[c]{real quality score \\matrix ($[0,1]^{n\times m}$)}&$\mathbb{I}$&\makecell[c]{collaborative self-\\information($\mathbb{R}^{n\times c}$)}\\ 
 	    \midrule
  	    $\mathcal{C}$ & \makecell[c]{classifier}&$\widetilde{\mathbb{I}}$&\makecell[c]{opposite collaborative \\ self-information}\\ 
  	    \midrule
 	    $\mathtt{E}_v$ & \makecell[c]{encoder for view $v$}&$\mathtt{D}_v$&\makecell[c]{decoder for view $v$}\\ 
 	    \midrule
   	    $\odot$ & \makecell[c]{Hadamard product}&$./$&\makecell[c]{the division of cor-\\responding elements}\\ 
   	    \midrule
   	     $\|\!\cdot\!\|_1$ & \makecell[c]{vector's $l_1$ norm}& $\|\!\cdot\!\|_2$&\makecell[c]{vector's $l_2$ norm}\\ 
	\bottomrule
	\end{tabular}}
\label{table:not}%
\end{table}
\section{Preliminary}
\subsection{Problem Formulation and Main Notations}
First, a given dataset includes $m$ views and $n$ samples can be denoted as: $\bigl\{\mathbf{X}^{(v)}\in \mathbb{R}^{n \times d_{v}}\bigr\}_{v=1}^{m}$ or $\{x_1^{(v)}, x_2^{(v)},...,x_n^{(v)}\}_{v=1}^{m}$, where $d_v$ is the feature dimension of $v$-th view. Any sample $i$ can be represented by $m$ feature vectors $\{x_i^{(1)}, x_i^{(2)},..., x_i^{(m)}\}$. And we define $\mathbf{Y} \in \bigl\{0,1\bigr\}^{n \times c}$ as the label matrix, where $c$ is the number of categories. $\mathbf{Y}_{i,j}=1$ indicates that the $i$-th sample is marked as the $j$-th category, otherwise $\mathbf{Y}_{i,j}=0$. Considering the compatibility for missing views and missing labels, we define two key prior matrices, the missing-view indicator $\mathbf{W} \in \bigl\{0,1\bigr\}^{n \times m}$ and the missing-label indicator $\mathbf{G} \in \bigl\{0,1\bigr\}^{n \times c}$. $\mathbf{W}_{i,j}=1$ and $\mathbf{G}_{i,j}=1$ mean $j$-th view and label of $i$-th sample is available respectively, and $\mathbf{W}_{i,j}=0$ or $\mathbf{G}_{i,j}=0$ is opposite. For simplicity, we fill random noise into the missing view in original data, and set '0' for unknown labels. Our goal is to train a neural network model on partially labeled incomplete multi-view data (each sample has its own label vector, although it may be incomplete) to perform inference on unlabeled test samples. To facilitate readers to quickly retrieve important notations, in Table \ref{table:not}, we list the main notations used in this paper. Besides, we make the following provisions on the representation of matrix subscripts: $\mathbf{A}_{i,j}$ indicates the element at $i$-th row and $j$-th column of matrix $\mathbf{A}$; $\mathbf{A}_{i,:}$ means $i$-th row vector of $\mathbf{A}$ and $\mathbf{A}_{:,j}$ is the $j$-th column vector.

\subsection{Contrastive Multi-view Learning}
\label{seclabel}
Contrastive learning is widely used in various unsupervised tasks, with the basic idea of minimizing the distance of positive sample-pairs while pushing away negative sample-pairs. Obviously, the effectiveness of this method directly depends on the construction or selection of positive and negative sample-pairs. A common version of the contrastive learning loss can be defined as follows \cite{oord2018representation,logeswaran2018efficient}: 
\begin{equation}
\label{eq:cl1}
\mathcal{L}_{c}=-\log \frac{e^{\mathcal{S}(x,x^+)}}{e^{\mathcal{S}(x,x^+)}+\sum_{x^-\in \mathcal{N}(x)}{e^{\mathcal{S}(x,x^-)}}},
\end{equation}
where $x,x^+,x^-$ denote the sample $x$ and its corresponding positive sample and negative sample. $\mathcal{N}(x)$ is a negative pairs set of $x$ and $\mathcal{S}$ is the cosine similarity function:
\begin{equation}
\mathcal{S}(x,y)= \frac{x^Ty}{\|x\|_2\cdot\|y\|_2}.
\end{equation}
The definition of positive and negative instance pairs in Eq. (\ref{eq:cl1}) is diversified under different applications. In multi-view contrastive learning, researchers align the representations of samples in different views to satisfy the multi-view consistency assumption \cite{xu2022multi,liu2023dicnet}. In these works, views of the same sample constitute positive pairs, and views of different samples are regarded as negative pairs. For instance, for arbitrary views $u$ and $v$ with $n$ samples, the similarity of positive pairs $\mathcal{S}(x,x^+)$ in Eq. (\ref{eq:cl1}) w.r.t. $i$-th sample is redefined as $\mathcal{S}(x_{i}^{(u)},x_{i}^{(v)})$, and that of negative pairs is $\{\mathcal{S}\big(x_{i}^{(u)},x_{j}^{(r)}\big)|j=1,...,n,j\neq i,r=u, v\}$. Observing these two definitions, we can easily find the so-called "negative pairs" are not \textit{truly} mutually exclusive since it is possible that samples $i$ and $j$ belong to the same cluster \cite{trosten2021reconsidering}. Although literature
\cite{trosten2021reconsidering} tries to assign only samples belong to different clusters as negative pairs through the clustering module, it is difficult to completely eliminate the class collision in this method, which relies heavily on clustering performance.

\subsection{Multi-view Weighted Fusion}
\label{secmwf}
A common strategy in feature fusion based multi-view learning methods is multi-view weighted fusion. To obtain a consistent representation of multiple views for outputting unique clustering or classification results, a simple but effective approach is to compute the average features of all views. However, many researchers believe that this approach defeats the original intention of the multi-view feature diversity, i.e., multiple views are of unequal importance \cite{wen2022survey}. Therefore, the weighted fusion technique is widely adopted in many existing works \cite{trosten2021reconsidering,liu2022localized,liu2023incomplete,wen2020dimc}. Assuming that $z_i^{(v)}$ is the learned representation of sample $i$ in view $v$, different weight coefficients are introduced for each view to obtain a weighted fusion representation:
\begin{equation}
\label{eq:wfusion}
\bar{z_i}=\sum_{v=1}^{m}a_vz_{i}^{(v)},
\end{equation}
$a_v$ is the positive weight of $v$-th view and $\sum_{v}^{m}a_v=1$.
However, these methods all consider the importance of a certain view to be invariant. In other words, overall high-quality views enjoy higher weight coefficients, which seems reasonable at first glance but ignores individual sample discrepancies. For instance, the RGB view of an image is easily classified into the correct category in well-lit conditions, but the infrared view generally exhibits stronger discriminative power in low-light conditions. Therefore, the '\textit{good}' views to different samples are not the same, and assigning the same weight to all instances in a view apparently violates the real-world laws. To the best of our knowledge, the vast majority of existing multi-view learning methods fail to tackle this issue.
\begin{figure*}[h!]
\centering
\includegraphics[width=0.99\textwidth]{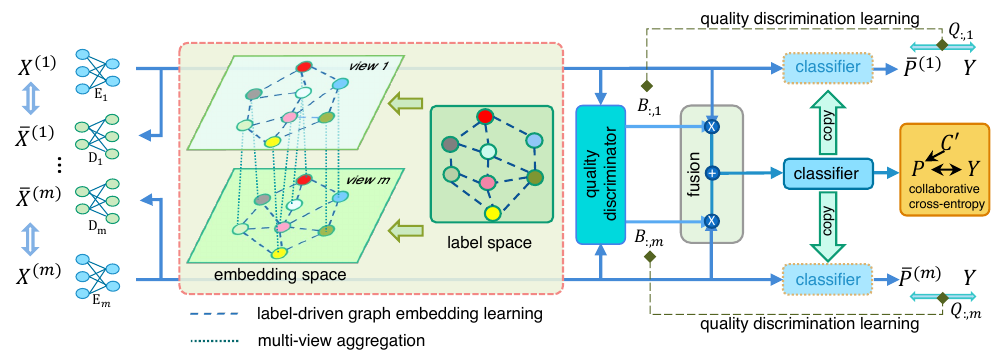} 
\caption{The main structure of our RANK. Each view is assigned a encoder and a decoder for feature extraction and data reconstruction. Multi-view aggregation and label-driven graph embedding learning are separately applied to the extracted features; the quality discriminator explicitly evaluates the instance-level view quality, providing dynamic weights for multi-view fusion; the final classification is performed by a collaborative cross-entropy classification module with label dependency awareness.}
\label{fig:main}
\end{figure*}
\subsection{Label Correlation Matrix}
\label{sec:lcm}
In existing works, the construction methods of the label correlation matrix are various, such as cosine similarity measurement based methods \cite{li2022learning,zhu2020global}, conditional probability based methods \cite{you2020cross,chen2019multi,ma2021label}, and some complex calculation rule based methods \cite{zhao2021consistency,wu2014multi}. In addition to explicitly constructing label correlation matrix according to prior information, some methods directly model a learnable label correlation matrix in their models \cite{zhao2021two,tan2018incomplete,tan2019individuality}. In this subsection, we specifically state an intuitive and simple method for generating conditional probability based label correlation matrix. For class labels $L_i$ and $L_j$, we model the dependence of $L_i$ on $L_j$ by the following conditional probability:
\begin{equation}
\label{eq:C}
\mathbf{C}_{i,j}=\mathcal{P}(L_j|L_i)=\frac{\sum_{k=1}^{n}\mathbf{Y}_{k,i}\mathbf{Y}_{k,j}}{\sum_{k=1}^{n}\mathbf{Y}_{k,i} }=\frac{\mathbf{Y}_{:,i}^T\mathbf{Y}_{:,j}}{\mathbf{Y}_{:,i}^T\mathbf{Y}_{:,i}},
\end{equation}
where $\mathbf{C} \in \mathbb{R}^{c\times c}$ is the label correlation matrix and $\mathcal{P}(L_j|L_i)$ denotes the probability of occurrence of label $L_j$ when label $L_i$ appears. Note that $\mathcal{P}(L_j|L_i) \neq \mathcal{P}(L_i|L_j)$ thus $\mathbf{C}$ is asymmetric. And the diagonal elements of $\mathbf{C}$ are all $1$. In other words, the $i$-th row of $\mathbf{C}$ indicates how dependent $L_i$ is on other labels, and the $i$-th column of $\mathbf{C}$ indicates this flipped dependency.

\section{Methodology}

\subsection{Feature Extraction and Reconstruction Framework}
Dissimilar to traditional methods that focus on and learn shallow-level features of data, we employ deep autoencoders commonly used in self-supervised learning methods to extract high-level embedding features of multi-view data. Specifically, each view is assigned a proprietary autoencoder network, consisting of an encoder and a decoder, where the encoder is the core unit of feature extraction and the decoder is responsible for reconstructing the original features. Such a seemingly simple design serves two purposes: (1) Aligning different views in the embedding space with same dimension to facilitate subsequent feature fusion. (2) Ensuring that the extracted high-level embedding features maintain the view-specific semantic information. The formal expression is: $\{\mathtt{E}_v(\mathbf{X}^{(v)})=\mathbf{Z}^{(v)}\}_{v=1}^{m}$ and $\{\mathtt{D}_v(\mathbf{Z}^{(v)})=\bar{\mathbf{X}}^{(v)}\}_{v=1}^m$, where $\mathtt{E}_v$ and $\mathtt{D}_v$ denote the encoder and decoder, which are composed of Multilayer Perceptrons (MLP), corresponding to $v$-th view. $\mathbf{Z}^{(v)} \in \mathbb{R}^{n\times d_e}$ is extracted embedding feature and $\bar{\mathbf{X}}^{(v)}\in \mathbb{R}^{n\times d_v}$ is the reconstructed original data of view $v$. Naturally, we minimize the reconstruction error by a weighted mean square error (MSE) loss:
\begin{equation}
\label{eq:lre}
\mathcal{L}_{re}=\frac{1}{m}\sum_{v=1}^{m}\bigg(\frac{1}{d_{v}n}\sum_{i=1}^{n}\Big\|x_{i}^{(v)}-\bar{x}^{(v)}_{i}\Big\|^{2}_{2}\mathbf{W}_{i,v}\bigg),
\end{equation}
where $\mathcal{L}_{re}$ is the reconstruction loss and $\bar{x}_i^{(v)}$ is corresponding reconstructed feature, i.e., $i$-th row of $\bar{\mathbf{X}}^{(v)}$. Unlike conventional MSE loss, a missing-view indicator matrix $\mathbf{W}$ is added to filter invalid views of each sample.
\subsection{Label-Driven Multi-view Contrastive Learning}\label{sec:l_d_cont}
Recalling the class conflicts that traditional unsupervised contrastive learning may cause in Section \ref{seclabel}, especially on the existing works in multi-view contrastive learning, an intuitive approach is to use the supervised information to select the truly negative pairs, i.e., only instances belonging to a different category are assigned to the object as paired negative instances, which is feasible in single-label classification. However, it is difficult for us to accurately judge whether two samples are absolutely positive or negative pairs based on multi-label supervision information, since their label vectors are likely to hold non-binary relevance. In addition, plenty of missing-tags make it harder to construct negative pairs, forcing us to change the existing contrastive multi-view learning strategy \cite{xu2022multi,trosten2021reconsidering}. Rethinking Eq. (\ref{eq:cl1}), it can be rewritten as following:
\begin{equation}
\label{eq:cl2}
\begin{aligned}
\mathcal{L}_{c}&=-\mathcal{S}(x,x^+)+\log\big(e^{\mathcal{S}(x,x^+)}+\sum_{x^-\in \mathcal{N}(x)}{e^{\mathcal{S}(x,x^-)}}\big)\\
&\approx\underbrace{-\mathcal{S}(x,x^+)}_{\text{Positive pair aggregation}}+\underbrace{\log\sum\limits_{x^-\in \mathcal{N}(x)}{e^{\mathcal{S}(x,x^-)}}}_{\text{Negative pair separation}}.
\end{aligned}
\end{equation}
\begin{figure}[t!]
\centering
\includegraphics[width=0.99\linewidth]{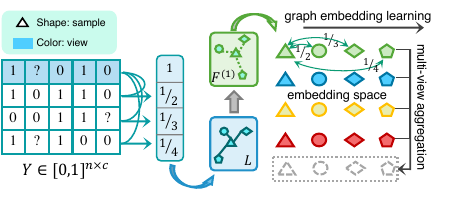} 
\caption{Label-driven multi-view contrastive learning. In the embedding space, we aggregate multiple views of the same sample across all view-planes and learn the geometric structure in the label space inside the view.}
\label{fig:contr}
\end{figure}
As shown in Eq. (\ref{eq:cl2}), the traditional contrastive learning method can be regarded as composed of positive pair aggregation and negative pair separation. In light of the above analysis on the selection of negative pairs, in this subsection, we only keep the positive pair aggregation term of Eq. (\ref{eq:cl2}) and replace the repelling of negative pairs with a label-driven structure-preserving method. To be specific, our label-driven multi-view contrastive learning consists of two parts, i.e., \textit{multi-view aggregation} and \textit{label-driven graph embedding learning}.
\subsubsection{Multi-view Aggregation}
Combined with the multi-view consensus assumption, it is easy to determine that multi-view features corresponding to the same sample are positive pairs of each other. Therefore, we retain the multi-view consensus aggregation in existing multi-view contrastive learning \cite{trosten2021reconsidering,xu2022multi,liu2023dicnet}, and obtain the optimization objective involving $i$-th sample's two views: 
\begin{equation}
\begin{aligned}
&\min-\mathcal{S}(z_i^{(u)},z_i^{(v)})\\
\Leftrightarrow&\min\big\|f\big(z_i^{(u)})-f\big(z_i^{(v)}\big)\big\|_2^2,
\end{aligned}
\end{equation}
where $z_i^{(u)}$ and $z_i^{(v)}$ respectively represent the $i$-th row of $\mathbf{Z}^{(u)}$ and  $\mathbf{Z}^{(v)}$ ($i$-th sample's embedding features on view $u$ and $v$). $f(x)$ denotes the $l_2$ normalization function. Then, the total multi-view aggregation loss can be expressed as:
\begin{equation}
\label{eq:lma}
\mathcal{L}_{ma}=\sum_{u=1}^{m}\sum_{\substack{v=1\\ v\neq u}}^{m}\frac{1}{N^{uv}d_e}\sum_{i=1}^{n} \mathbbmss{1}_{[\Psi]}\big\|f\big(z_i^{(u)})-f\big(z_i^{(v)}\big)\big\|_2^2,
\end{equation}
where $\mathbbmss{1}_{[\Psi]}$ is equal to $1$ when the condition $\{\Psi: \mathbf{W}_{i,u}\mathbf{W}_{i,v}=1\}$ is met, otherwise 0. This condition is to ensure that only when two views are available at the same time, their errors are calculated into the loss. Accordingly, $N^{uv} = \|\mathbf{W}_{:,u}\odot \mathbf{W}_{:,v}\|_1$ represents the number of samples that satisfy the condition.

\subsubsection{Label-Driven Graph Embedding Learning}

As mentioned above, it is difficult to use binary values to define whether two samples are close or far based on multi-label information, but we have the opportunity to utilize the non-equidistant property of the multi-label space (the Euclidean distance between samples does not have to be equal to $\sqrt{2}$ or 0 like in the single-label space) to obtain a strongly supervised geometric structure information among samples. Specifically, we define following approach to calculate the similarity matrix among samples in the incomplete multi-label space:  
\begin{equation}
\label{eq:T}
\mathbf{L}=[({\mathbf{Y}}{\mathbf{Y}}^T)./({\mathbf{G}}{\mathbf{G}}^T)]_{r\_max\_norm},
\end{equation}
where $\mathbf{L} \in [0,1]^{n\times n}$ is the similarity matrix of $n$ samples in label space and $[\cdot]_{r\_max\_norm}$ means the row-maximum normalization. We introduce missing-label indicator matrix $G$ to help reduce the negative effects of missing tags. And the purpose of $\mathbf{G}\mathbf{G}^T$ is to normalize the similarity value, whose element $(\mathbf{G}\mathbf{G}^T)_{i,j}$ is the number of labels jointly available for samples $i$ and $j$. Such an $\mathbf{L}$ can be regarded as a label-driven graph with strongly supervised information, which is very different from the affinity graph in most existing methods \cite{liu2022localized,wen2019unified,wen2022survey}, constructed by Gaussian kernel function and KNN in original feature space. And with such a prior graph, we can push the geometric structure of samples in the embedding space to be consistent with that in the label space, i.e., label-driven feature structure preservation. To achieve this, we introduce a global similarity graph $\mathbf{F}^{(v)}$ from embedding feature space, whose element is calculated by:
\begin{equation}
\mathbf{F}_{ij}^{(v)}=(\mathcal{S}(z_i^{(v)},z_j^{(v)})+1)/2,
\end{equation}
where $\mathbf{F}^{(v)} \in [0,1]^{n\times n}$ and we can get $\{\mathbf{F}^{(v)}\}_{v=1}^{m}$ for $m$ views. Note that our $\mathbf{F}^{(v)}$ and $\mathbf{L}$ are both globally aware, describing how similar each sample is to all other samples. Next, we regard $\mathbf{L}$ as the target graph and use the cross-entropy loss to guide the learning of $\mathbf{F}^{(v)}$:
\begin{equation}
\label{eq:lge}
\begin{aligned}
\mathcal{L}_{ge} &= -\frac{1}{2mN^{ij}}\sum_{v=1}^{m}\sum_{i=1}^{n}\sum_{j\ne i}^{n}\mathbbmss{1}_{[\Upsilon]}\bigl({\mathbf{L}}_{ij}\log{{\mathbf{F}}^{(v)}_{ij}}\\
&+\bigl(1-{\mathbf{L}}_{ij}\bigr)\log{\bigl(1-{\mathbf{F}}^{(v)}_{ij}\bigr)}\bigr),
\end{aligned}
\end{equation}
where $\mathbbmss{1}_{[\Upsilon]} = 1$ if condition $\{\Upsilon: \mathbf{W}_{iv}\mathbf{W}_{jv}=1\}$ is true, 0 otherwise. $\mathbbmss{1}_{[\Upsilon]}$ aims to masks out the combinations with any missing samples.  And $N^{ij}=\sum_{i,j}{{\mathbf{W}}_{iv}{\mathbf{W}}_{jv}}$ represents the number of valid sample combinations.

Note that due to the memory limitations, it is common to train the model in the mini-batch manner, so calculating the similarity graph within a mini-batch of the data may lead to a local action range of the constraint. To cope with this, in the training process, we store the embedding features of all instances at each training epoch, and then calculate the similarity graph form current batch to entire training data. As a result, we can get a global similarity graph $F\in [0,1]^{n_b\times n_a}$ rather than a partial similarity graph $F\in [0,1]^{n_b\times n_b}$ in the mini-batch training, where $n_b$ and $n_a$ are batch size and entire training size, respectively. It should be noted that the stored global embedding features of $n_a$ samples will be updated immediately as soon as the new embedding features of mini-batch data are encoded at the current iteration. Since at the beginning of training, we cannot obtain the global features of the last epoch, the loss $\mathcal{L}_{ge}$ will be executed from the second epoch.

\begin{figure}[t!]
\centering
\includegraphics[width=0.99\linewidth]{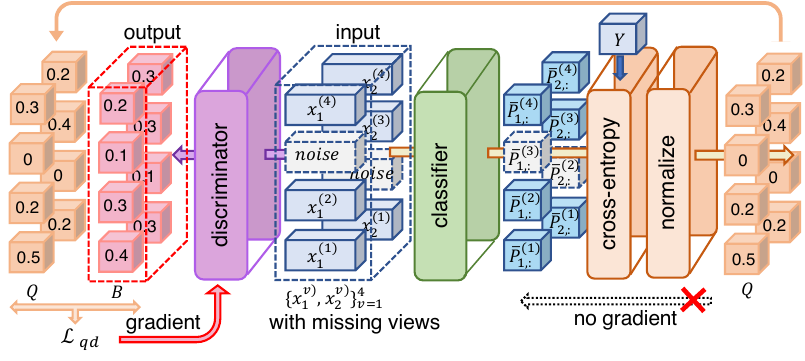} 
\caption{The schematic diagram of the quality discrimination learning. The parameter update of the quality discriminator is only driven by the loss $\mathcal{L}_{qd}$ on the left.}
\label{fig:lqd}
\end{figure}
\subsection{Quality Discrimination Learning and Dynamically Weighted Fusion}
As mentioned in Section \ref{secmwf}, roughly assigning a fixed weight to each view to balance the view relationship is too rigid to be usually satisfied in practice. We expect the network to actively judge the views' quality of each sample, and then dynamically assign different weights to different views of the sample. To achieve this, in this subsection, we propose a novel quality-aware sub-network to score all views of each sample. The architecture of the quality discriminator is very simple, consisting of two fully connected (FC) layers, an activation layer, and a softmax function, which is fed $m$ views' embedding features and outputs corresponding quality scores: $\mathcal{D}(\{\mathbf{Z}^{(v)}\}_{v=1}^{m})= \mathbf{B}$, where $\mathcal{D}$ is the discriminator. $\mathbf{B}\in [0,1]^{n\times m}$ denotes the entire quality scores (weight matrix) and the views' scores belonging to one sample sum to 1. Obviously, the feed-forward process of the discriminator is very simple but the success of it hinges on a good design of the learning objective, that is, how to supervise the learning of the quality discriminator effectively.

Going back to an essential and intuitive hypothesis, if a certain view's quality of a sample is good, then its feature is more likely to be classified into the correct categories. In view of this, we directly compute $m$ view-specific classification results through the classifier of the main network: $\mathcal{C}(\{\mathbf{Z}^{(v)}\}_{v=1}^{m})=\{\bar{\mathbf{P}}^{(v)}\}_{v=1}^{m}$, where $\mathcal{C}$ is our classifier trained in the main network and $\bar{\mathbf{P}}^{(v)}\in [0,1]^{n\times c}$ is $v$-th view's prediction. With the $m$ view-specific predicted label matrices, we can measure the true view quality using the cross-entropy loss function:
\begin{equation}
\begin{aligned}
\mathbf{Q}'_{i,v} &= \frac{1}{\|\mathbf{G}_{i,:}\|_1}\sum\limits_{j=1}^{c}\Big({\mathbf{Y}}_{i,j}\log\bar{\mathbf{P}}^{(v)}_{i,j}\\
&+(1-\mathbf{Y}_{i,j})\log(1-\bar{\mathbf{P}}^{(v)}_{i,j})
\Big){\mathbf{G}}_{ij},
\end{aligned}
\end{equation}
where $\mathbf{Q}'_{i,v}$ is a direct quality score of $v$-th view of $i$-th sample (less than 0). The larger the value, the smaller the prediction error, and the higher the quality  of  $v$-th view. Further, we normalize $\mathbf{Q}'$ to be consistent with the quality score matrix $\mathbf{B}$ from the discriminator $\mathcal{D}$ by:
\begin{equation}
\mathbf{Q}_{i,v} = \frac{\exp({\mathbf{Q}'_{i,v}})\mathbf{W}_{i,v}}{\sum_{v}^{m}\exp(\mathbf{Q}'_{i,v})\mathbf{W}_{i,v}}.
\end{equation}
$\mathbf{Q} \in [0,1]^{n\times m}$ is regarded as the target score matrix of $\mathcal{D}$. Calculate the cross-entropy between $\mathbf{Q}$ and $\mathbf{B}$ to get the final quality discrimination loss $\mathcal{L}_{qd}$:
\begin{equation}
\label{eq:lqd}
\mathcal{L}_{qd}= -\frac{1}{n}\sum_{i=1}^{n}\sum_{j=1}^{m}\mathbf{Q}_{i,j}\log\mathbf{B}_{i,j}.
\end{equation} 

Finally, these quality scores will be used in the dynamically weighted fusion module to obtain high-quality multi-view fusion feature:
\begin{equation}
\label{eq:dwf}
\bar{\mathbf{Z}}=\sum_{v=1}^{m}diag(\mathbf{B}_{:,v})\mathbf{Z}^{(v)},
\end{equation}
where $\bar{\mathbf{Z}}\in \mathbb{R}^{n\times d_e}$ is the multi-view fusion feature that is used for final classification. And $diag(\mathbf{B}_{:,v})$ means the diagonal matrix with diagonal $\mathbf{B}_{:,v}$.

Note that the view quality scores output to the main network do \textit{not} participate in the gradient calculation of the main network, because we require the learning and decision-making of the discriminator to be relatively independent of the fusion and classification process of the main network, that is, the $\mathcal{D}$ is not subject to the main classification loss (see below subsection).
\begin{figure}[t!]
\centering
\includegraphics[width=0.99\linewidth]{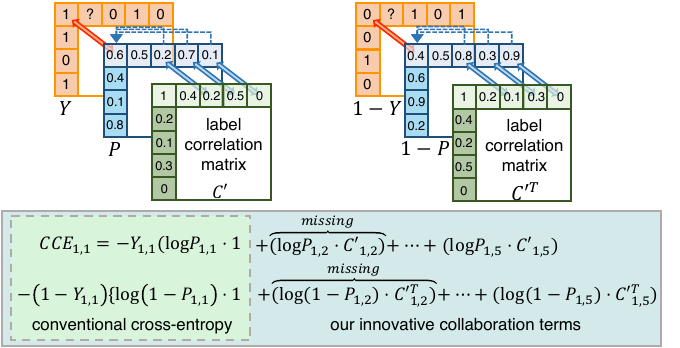} 
\caption{Schematic diagram of our multi-label collaborative cross-entropy loss $\mathcal{L}_{mcce}$. The ${CCE}_{1,1}$ represents the collaborative cross-entropy of the first sample in the first category.}
\label{fig:mcc}
\end{figure}
\subsection{Multi-label Collaborative Classification}
After the aforementioned feature extraction network and dynamically weighted fusion module, we get the multi-view fusion representation $\bar{\mathbf{Z}}$ for the final classification purpose. Briefly, we use a classifier consisting of an FC layer and a Sigmoid activation function to obtain the final predicted result: $\mathcal{C}(\mathbf{\bar{Z}})=\mathbf{P} \in [0,1]^{n\times c}$. To guide the learning of the main network via the real label $\mathbf{Y}$, in the existing deep multi-label classification methods, the multi-label binary cross-entropy (BCE) loss $\mathcal{L}_{mbce}$ is generally adopted to calculate the prediction error \cite{ma2021label,chen2019multi,lyu2022beyond}:
\begin{equation}
\label{eq:mbce}
\begin{aligned}
{\mathcal{L}}_{mbce} &=  -\frac{1}{nc}\sum\limits_{i=1}^{n}\sum\limits_{j=1}^{c}\Big({\mathbf{Y}}_{ij}\log({\mathbf{P}}_{ij})\\
&+(1-{\mathbf{Y}}_{ij})\log(1-{\mathbf{P}}_{ij})
\Big).
\end{aligned}
\end{equation}
Observing Eq. (\ref{eq:mbce}), we find that it only repeatedly performs the BCE loss, commonly used in binary classification tasks, on multiple labels. That is to say, this multi-label BCE loss considers the multi-label classification problem as multiple independent binary classification problems without any label correlation. To solve this issue, we innovatively propose a multi-label collaborative cross-entropy loss. As we know, when label $L_j$ appears, labels $\{L_1, L_2, L_3, ...,L_{j-1}, L_{j+1},..., L_c\}$ appear with corresponding probabilities. First, we need a correlation matrix that can be calculated using Eq. (\ref{eq:C}). And then, we smooth it out by:
\begin{equation}
\label{eq:C1}
\mathbf{C}'_{i,j}=\bigg\{
\begin{aligned}
&\mathbf{C}_{i,j},& \text{if }\mathbf{C}_{i,j}>\sigma\\
&0, & \text{otherwise}
\end{aligned},
\end{equation}
$\sigma\in [0,1]$ is the truncation threshold and $\mathbf{C}'$ is the smooth version of $\mathbf{C}$. Extending this dependence to the calculation of information entropy \cite{shannon1948mathematical}, we can formulate  $i$-th sample' collaborative self-information $\mathbb{I}_{i,j}$ w.r.t. label $L_j$:
\begin{equation}
\label{eq:ce1}
{\mathbb{I}_{i,j}}=\underbrace{\mathbf{C}'_{j,1}\mathbf{I}_{i,1}\mathbf{G}_{i,1}+\mathbf{C}'_{j,2}\mathbf{I}_{i,2}\mathbf{G}_{i,2}+...+\mathbf{C}'_{j,c}\mathbf{I}_{i,c}\mathbf{G}_{i,c}}_{\text{Sum of }c \text{ terms}},
\end{equation}
where $\mathbf{I}_{i,j}=-\log\mathbf{P}_{i,j}$ represent the self-information of $\mathbf{P}_{i,j}$. In contrast, we consider the effect of sample $i$ not being tagged as $L_j$ on other labels and formulate the opposite $\widetilde{\mathbb{I}}_{i,j}$ as:
\begin{equation}
\label{eq:ce2}
\widetilde{\mathbb{I}}_{i,j}=\mathbf{C}'_{1,j}\widetilde{\mathbf{I}}_{i,1}\mathbf{G}_{i,1}+\mathbf{C}'_{2,j}\widetilde{\mathbf{I}}_{i,2}\mathbf{G}_{i,2}
+...+\mathbf{C}'_{c,j}\widetilde{\mathbf{I}}_{i,c}\mathbf{G}_{i,c},
\end{equation}
where $\widetilde{\mathbf{I}}_{i,j} =-\log(1-\mathbf{P}_{i,j})$. As can be clearly seen from Eq. (\ref{eq:ce1}) and Eq. (\ref{eq:ce2}), the calculation of each sample's collaborative self-information includes that of all samples related to it. In other words, they can be regarded as an information aggregation strategy based on label dependencies. $\mathbf{G}$ is introduced to skip the invalid label. Combining Eq. (\ref{eq:ce1}) and Eq. (\ref{eq:ce2}), our multi-label collaborative cross-entropy loss $\mathcal{L}_{mcce}$ can be formalized as:
\begin{equation}
\label{eq:lmcce}
\mathcal{L}_{mcce}=\frac{1}{n}\sum_{i=1}^{n}\sum_{j=1}^{c}(\mathbf{Y}_{i,j}\mathbb{I}_{i,j}+(1-\mathbf{Y}_{i,j})\widetilde{\mathbb{I}}_{i,j})\mathbf{G}_{i,j}.
\end{equation}
At first glance, Eq. (\ref{eq:lmcce}) is similar in form to Eq. (\ref{eq:mbce}) but has different connotations. Our $\mathcal{L}_{mcce}$ cleverly combines the multi-label correlation information to break the inter-class barriers in conventional multi-label cross-entropy. Besides, it is easily migrated to other deep incomplete multi-label classification networks.

\subsection{Overview of the Method}
\subsubsection{Overall Objective Function}
Combining the 5 loss functions of the above 4 parts, we can get the following overall objective function:
\begin{equation}
\label{eq:lall}
\mathcal{L}_{all} = \gamma\mathcal{L}_{re}+(1-\beta^{t})\mathcal{L}_{ma}+\alpha\mathcal{L}_{ge}+\mathcal{L}_{qd}+\mathcal{L}_{mcce},
\end{equation}
where $t$ is the current training epoch, and $\alpha$, $\beta$, and $\gamma$ are penalty parameters. We set an incremental coefficient for $\mathcal{L}_{ma}$ for this consideration: At the beginning of training, the features extracted by encoders are still accompanied by more noise, without strong discrimination and semantics. At the moment, rigidly aligning all views is meaningless or even harmful. When quality discrimination learning and graph embedding learning reach stability in the middle and late stages of training, the direction of consensus aggregation will shift to high-quality views (See Sections \ref{sec:abe} and \ref{sec:ps} for more analysis). In algorithm \ref{al:1}, we show the detailed training process.
\begin{algorithm}[!t]
	\caption{Training process of \textbf{RANK}}
	\label{al:1}

	\begin{algorithmic}[1]
	\Require Incomplete multi-view data $\big\{X^{(v)}\big\}_{v=1}^{m}$, missing-view indicator matrix $\mathbf{W}$, incomplete label matrix $\mathbf{Y}$, missing-label indicator matrix $\mathbf{G}$.
	\Ensure Prediction $\mathbf{P}$.
	\State Compute target graph $\mathbf{L}$ by Eq. (\ref{eq:T}).
	\State Initialize model parameters and set hyperparameters ($\alpha$, $\beta$, $\gamma$, learning rate, batch size, and training epochs $e$).
	\State Compute the global label correlation graph $\mathbf{C}'$ by Eq. (\ref{eq:C}) and Eq. (\ref{eq:C1}).
	\State t=0.
		\While{$t< e$}                                                	
		\State Extract multi-view embedding feature ${\{\mathbf{Z}^{(v)}\}_{v=1}^{m}}$ by encoders $\{\mathtt{E}_v\}_{v=1}^{m}$.
		\State Reconstruct data $\big\{\bar{\mathbf{X}}^{(v)}\big\}_{v=1}^{m}$ by decoders $\{\mathtt{D}_v\}_{v=1}^{m}$ and compute reconstruction loss $\mathcal{L}_{re}$ by Eq. (\ref{eq:lre}).
		\State Compute multi-view aggregation loss using Eq. (\ref{eq:lma}).
		\If{$t>0$}
		\State Compute similarity graph $\mathbf{F}$ in embedding space with ${\{\mathbf{Z}^{(v)}\}_{v=1}^{m}}$ saved in last epoch.
		\State Compute loss $\mathcal{L}_{ge}$ using Eq. (\ref{eq:lge}).
		\EndIf
		\State Input ${\{\mathbf{Z}^{(v)}\}_{v=1}^{m}}$ into discriminator $\mathcal{D}$ and get quality scores matrix $\mathbf{B}$.
		\State Compute quality discrimination loss $\mathcal{L}_{qd}$ by Eq. (\ref{eq:lqd}).
		\State Obtain weighted fusion representation $\bar{\mathbf{Z}}$ according to Eq. (\ref{eq:dwf})
		\State Compute predicted result $\mathbf{P}$ by classifier $\mathcal{C}$.
		\State Compute multi-label collaboration classification loss $\mathcal{L}_{mcce}$ using Eq. (\ref{eq:lmcce}).
		\State Compute total loss $\mathcal{L}_{all}$ by Eq. (\ref{eq:lall}) and update network parameters.
		\State $t = t+1$.
		\State Save ${\{\mathbf{Z}^{(v)}\}_{v=1}^{m}}$.
		\EndWhile
	\end{algorithmic}
	\vspace{-0.1cm}
\end{algorithm}
\begin{figure*}[!t]
\centering
\subfloat[Corel5k database with 50\% missing labels]{
\includegraphics[width=0.4\textwidth]{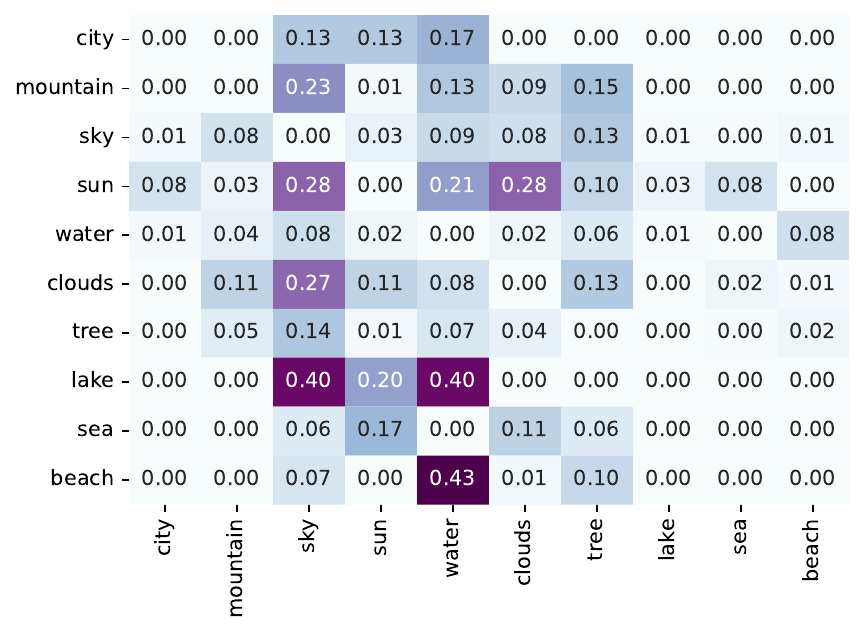}
}
\subfloat[Corel5k database with full labels]{
\includegraphics[width=0.4\textwidth]{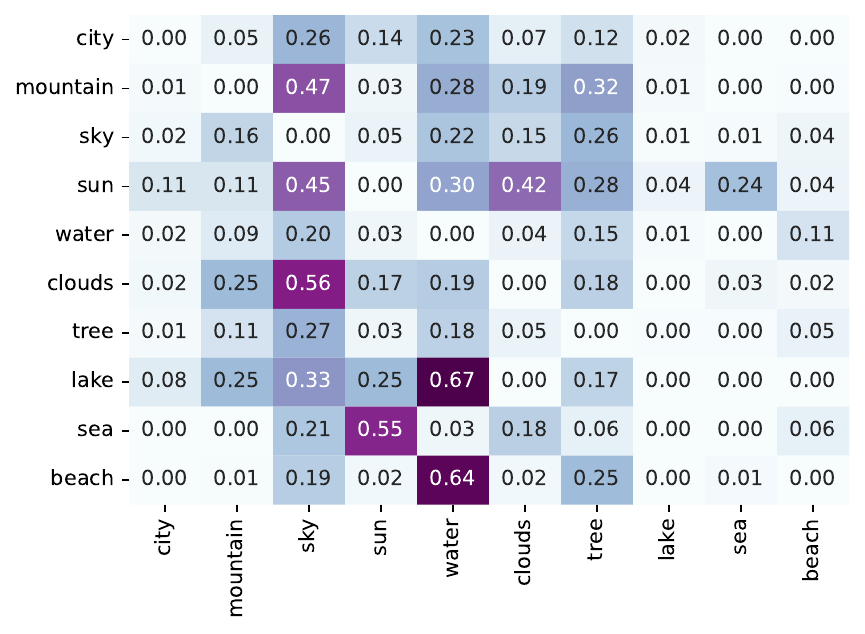}
} 
\quad
\subfloat[Pascal07 database with 50\% missing labels]{
\includegraphics[width=0.4\textwidth]{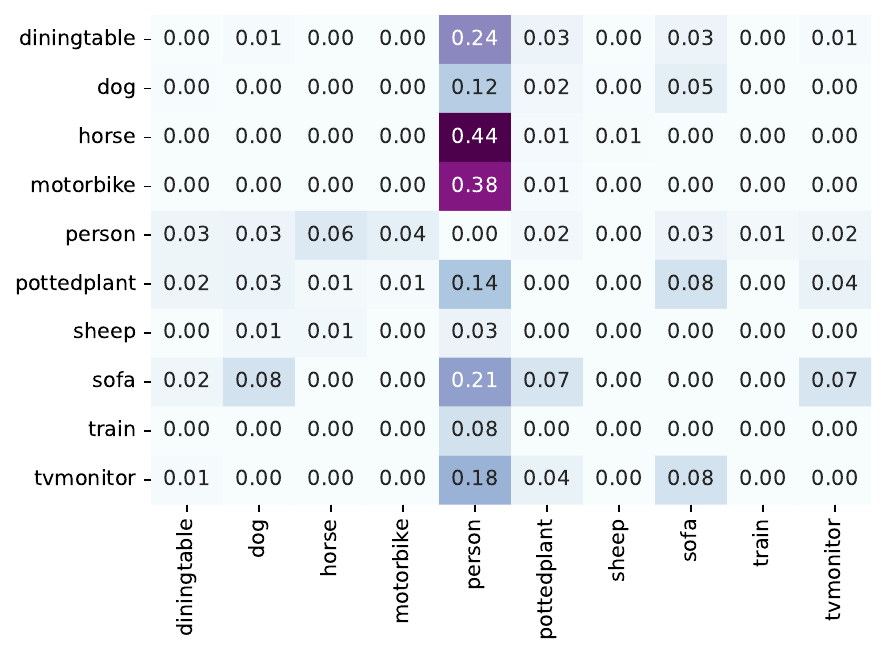}
}
\subfloat[Pascal07 database with full labels]{
\includegraphics[width=0.4\textwidth]{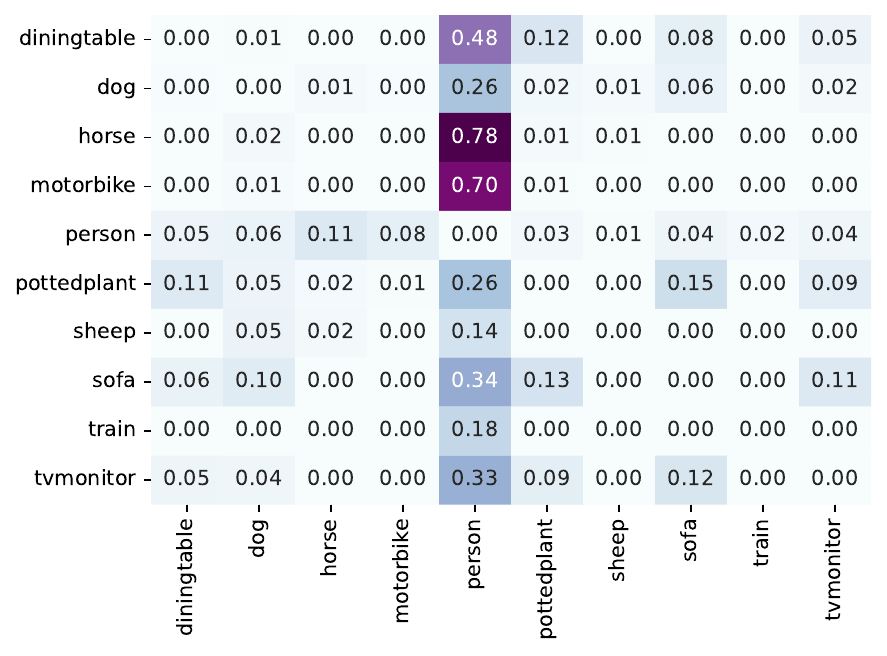}
} 
\caption{The heat maps of label correlation matrices for ten categories on Corel5k and Pascal07 dataset with (a,c) 50\% missing labels and (b,d) full labels.}
\label{fig:lrm}
\end{figure*}
\subsubsection{Complexity Analysis}
Although we design 5 loss functions for our RANK, its main computational overhead still comes from the simple autoencoder network, and next is our specific analysis. For convenience, we add that the maximum number of neurons in the middle layers of the network is $d_{max}$, and the time complexity of the main autoencoder is $O(nmd_{max}^2)$. In addition, the time complexity of the five loss functions is $O(mn)$, $O(m^2n)$, $O(mn^2)$, $O(mnc)$, and $O(nc^2)$, corresponding to $\mathcal{L}_{re}$, $\mathcal{L}_{ma}$, $\mathcal{L}_{ge}$, $\mathcal{L}_{qd}$, and $\mathcal{L}_{mcce}$, respectively. The overall time complexity is $O(nmd_{max}^2+mn+m^2n+mn^2+mnc+nc^2)$. The complexity of our RANK is proportional to the square of various sizes of the dataset, such as the feature dimension, the number of samples, the number of views, and the number of categories, covering almost all key attributes of multi-view multi-label datasets. This also reflects that our RANK deeply mines information of the dataset from various aspects.

\subsubsection{Some Remarks}
In this subsection, we give some remarks on technical details and potential doubts to facilitate readers' understanding of our method.
\begin{remark}
\textbf{What is the difference between Eq. (\ref{eq:lge}) and existing graph learning methods?} 
Eq. (\ref{eq:lge}), which is formally similar to the multi-label cross-entropy classification loss, is actually an "inverse consistency graph learning" module. Unlike most graph learning based methods that learn a high-quality consensus graph from multi-view features for downstream clustering or classification, our method utilizes a common graph with strong supervision information to guide the individual views' graph learning, the essence of which is still to learn reliable embedding representations.
\end{remark}
\begin{remark}
\textbf{Why not introduce the missing-view indicator matrix in Eq. (\ref{eq:dwf}) like other deep incomplete multi-view learning methods \cite{liu2023dicnet,wen2022survey}}? In Eq. (\ref{eq:dwf}), we dynamically weight and fuse all views without prior incompleteness information, thanks to our quality discriminator being able to accurately distinguish invalid noisy data and assign them weights close to 0. In other words, the strong view discriminative information output by the quality discriminator covers the weak view discriminative information contained in the missing-view indicator.
\end{remark}
\begin{remark}
\textbf{The impcat of missing labels on the construction of the label correlation matrix.}
Of course, missing labels will inevitably damage the validity of the label correlation matrix, however, we found through statistics and experiments that the approximate label correlation matrix obtained based on the missing label matrix, is basically consistent with its complete version in structure. As shown in Fig. (\ref{fig:lrm}), we draw the heat maps of the sub-correlation matrices containing the 10 labels on the Corel5k and Pascal07 datasets with 50\% missing labels and full labels respectively. It can be seen that although the approximate label correlation matrix is numerically different from the complete version, the dependency relationship and main structure among labels do not change greatly, which is due to the zero-filling strategy for missing labels and the adopted construction method of label correlation matrix.
\end{remark}
\begin{remark}
\textbf{Why introduce a truncation threshold $\sigma$ in Eq. (\ref{eq:C1})?} Observing Eq. (\ref{eq:ce1}), we can find that the elements of the correlation matrix can be regarded as the weights of the self-information. We expect that the self-information terms with high weights have a greater impact on the calculation of collaborative self-information, and terms with low weights have a slight impact. However, an $\mathbf{I}_{i,j}$ with a low weight have opportunity to break this ideal assumption as the prediction $\mathbf{P}_{i,j}$ tends to zero. Therefore, we introduce a truncation threshold $\sigma$ to let the overall collaborative self-information pay more attention to high-relevance labels and ignore terms corresponding to low-relevance labels. In addition, the selection of $\sigma$ also depends on the label missing rate, we recommend a lower threshold for datasets with high label missing rate as the label correlation calculated by missing labels is usually numerically lower than that calculated by full labels, which can be found in Fig. (\ref{fig:lrm}).
\end{remark}

\section{Experiments}
In this section, we introduce the dataset settings, comparison methods and analyze the experimental results in detail. 

\subsection{Experimental Settings}
\begin{table}[t]
	\centering
	\caption{Detailed statistics about five multi-view multi-label databases.}
	\label{tab:data}
	\resizebox{0.49\textwidth}{!}{
		\begin{tabular}{ccccc}
			\toprule[1.2pt]
			Database   & \# Sample  & \# Label & \# View &   \# Label/\#Sample       \\ \midrule
			Corel5k &    4999    &    260    &    6    &    3.40    \\
			Pascal07   &    9963     &    20    &    6    & 1.47 \\
			ESPGame  &    20770     &    268    &    6     &     4.69     \\
			IAPRTC12   &    19627     &    291    &    6     &  5.72   \\
			MIRFLICKR   &    25000     &    38    &    6    &     4.72     \\ 
			\bottomrule[1.2pt]
	\end{tabular}}
\end{table}

\subsubsection{Datasets}
Following existing works \cite{tan2018incomplete,li2022concise,liu2023dicnet}, we select five popular multi-view multi-label datasets to evaluate our method, namely Corel5k, Pascal07, ESPGame, IAPRTC12, and Mirflickr databases. All of these datasets are uniformly extracted six kinds of feature as six views, i.e., GIST, HSV, Hue, Sift, RGB, and LAB. Table \ref{tab:data} shows the basic information of the five datasets and we introduce them in detail below:
\begin{enumerate}
\item \textbf{Corel5k} \cite{duygulu2002object}: Corel5k dataset is composed of 4999 images with 260 annotations, each image is labeled with 1 to 5 tags.
\item \textbf{Pascal07} \cite{everingham2009pascal}:  Pascal voc 2007 is a wildly used image dataset in the field of visual object detection and recognition. In our experiments, 9963 images and 20 kinds of objects are used.
\item \textbf{ESPGame} \cite{von2004labeling}: ESPGame dataset consists of 20770 images taken from online interactive games and 1 to 15 labels are extracted from the game for each image. The ESPGame contains an average of 4.69 semantic labels per image, with a total of 268 label annotations.
\item \textbf{IAPRTC12} \cite{henning2006iapr}: IAPRTC12 is also a large-scale dataset with 19627 image samples and 291 categories. Each sample contains up to 23 labels, which are extracted from the slogan or subtitles in the image.
\item \textbf{Mirflickr} \cite{huiskes2008mir}: Mirflickr-25k open evaluation project is composed of 25000 images downloaded from the social photography site Flickr. 38 labels are used in the experiments.
\end{enumerate}
In some tables and figures, the abbreviations of these datasets will be adopted to save space, i.e., COR, PAS, ESP, IAP, and MIR.
\subsubsection{Incomplete Dataset Preparation}
Since the above five datasets are all with complete views and labels, similar to existing works \cite{tan2018incomplete,li2022concise,liu2023dicnet}, we manually prepare incomplete multi-view missing multi-label datasets to evaluate the compatibility of our method to incomplete cases. To be more specific, 50\% of instances on each view are randomly removed while ensuring that the total number of samples does not decrease, i.e., each sample has at least one available view. Then, we set 50\% of the positive and negative labels for each category to be unknown. Finally, we randomly partition the entire dataset into training set (70\%), validation set (15\%), and test set (15\%). Note that the features of entire dataset including the validation and test sets are incomplete, but the labels of the test and validation sets are complete to accurately evaluate the performance.
\begin{table}[t]
	\centering
	\caption{Comparison multi-label classification methods. 'Multi-view' means the method is designed for multi-view data; 'Missing-view' and 'Missing-label' denote the method is able to handle incomplete views and missing labels respectively.}
	\label{tab:methods}
	\resizebox{0.49\textwidth}{!}{
		\begin{tabular}{ccccc}
			\toprule[1.2pt]
			Method   & Sources & Multi-view  & Missing-view & Missing-label \\ \midrule
			C2AE &    AAAI '17    &    \XSolidBrush    &    \XSolidBrush    &    \Checkmark    \\
			GLOCAL &    TKDE '17    &    \XSolidBrush    &    \XSolidBrush    &    \Checkmark    \\
			CDMM &    KBS '20    &    \Checkmark    &    \XSolidBrush    &    \XSolidBrush    \\
			DM2L &    PR '21    &    \XSolidBrush    &    \XSolidBrush    &    \Checkmark    \\
			LVSL &    TMM '22    &    \Checkmark    &    \XSolidBrush    &    \XSolidBrush    \\
			iMVWL &    IJCAI '18    &    \Checkmark    &    \Checkmark    &    \Checkmark    \\
			NAIM3L &    TPAMI '22    &    \Checkmark    &    \Checkmark    &    \Checkmark    \\
			DICNet &    AAAI '23    &    \Checkmark    &    \Checkmark    &    \Checkmark    \\
			\bottomrule[1.2pt]
	\end{tabular}}
\end{table}

\begin{table*}[t!]
\renewcommand{\arraystretch}{1.1}
\centering
\sc
\caption{Experimental results of nine methods on the five datasets with 50\% missing instances, 50\% missing labels, and 70\% training samples. $\uparrow$ indicates that the higher the value, the better the performance, and $\downarrow$ is the opposite. 'Ave.R' represents the average ranking of the corresponding method on all six metrics.}
\label{table:inc}
\small
    \begin{tabular}{p{0.7cm}p{0.95cm}cccccccc>{\columncolor{Gray}}c}
	\toprule
    Data & Metric & C2AE   & GLOCAL    & CDMM  &DM2L &LVSL &iMVWL&NAIM3L&DICNet&RANK\\
    \midrule	
    \multirow{6}[3]{*}{\begin{turn}{0}{COR}\end{turn}}
               &AP$\uparrow$	&$\mathsf{0.227_{0.008}}$	&$\mathsf{0.285_{0.004}}$	&$\mathsf{0.354_{0.004}}$	&$\mathsf{0.262_{0.005}}$	&$\mathsf{0.342_{0.004}}$	&$\mathsf{0.283_{0.008}}$	&$\mathsf{0.309_{0.004}}$	&$\mathsf{0.381_{0.004}}$	&$\mathsf{0.427_{0.006}}$\\
               &1-HL$\uparrow$	&$\mathsf{0.98_{0.002}}$	&$\mathsf{0.987_{0.000}}$	&$\mathsf{0.987_{0.000}}$	&$\mathsf{0.987_{0.000}}$	&$\mathsf{0.987_{0.000}}$	&$\mathsf{0.978_{0.000}}$	&$\mathsf{0.987_{0.000}}$	&$\mathsf{0.988_{0.000}}$	&$\mathsf{0.988_{0.000}}$\\
               &1-RL$\uparrow$	&$\mathsf{0.804_{0.01}}$	&$\mathsf{0.84_{0.003}}$	&$\mathsf{0.884_{0.003}}$	&$\mathsf{0.843_{0.002}}$	&$\mathsf{0.881_{0.003}}$	&$\mathsf{0.865_{0.005}}$	&$\mathsf{0.878_{0.002}}$	&$\mathsf{0.882_{0.004}}$	&$\mathsf{0.913_{0.002}}$\\
               &AUC$\uparrow$	&$\mathsf{0.806_{0.01}}$	&$\mathsf{0.843_{0.003}}$	&$\mathsf{0.888_{0.003}}$	&$\mathsf{0.845_{0.002}}$	&$\mathsf{0.884_{0.003}}$	&$\mathsf{0.868_{0.005}}$	&$\mathsf{0.881_{0.002}}$	&$\mathsf{0.884_{0.004}}$	&$\mathsf{0.916_{0.003}}$\\
               &OE$\downarrow$	&$\mathsf{0.754_{0.016}}$	&$\mathsf{0.673_{0.01}}$	&$\mathsf{0.59_{0.007}}$	&$\mathsf{0.705_{0.014}}$	&$\mathsf{0.609_{0.009}}$	&$\mathsf{0.689_{0.015}}$	&$\mathsf{0.65_{0.009}}$	&$\mathsf{0.532_{0.007}}$	&$\mathsf{0.501_{0.012}}$\\
               &Cov$\downarrow$	&$\mathsf{0.404_{0.016}}$	&$\mathsf{0.352_{0.006}}$	&$\mathsf{0.277_{0.007}}$	&$\mathsf{0.353_{0.005}}$	&$\mathsf{0.282_{0.006}}$	&$\mathsf{0.298_{0.008}}$	&$\mathsf{0.275_{0.005}}$	&$\mathsf{0.273_{0.011}}$	&$\mathsf{0.200_{0.005}}$\\
               \rowcolor{LGray}&Ave.R$\downarrow$&8.83 	&6.33 	&2.83 	&6.83 	&3.83 	&6.83 	&4.33 	&2.17 	&1.00 \\
    \midrule
    \multirow{6}[3]{*}{\begin{turn}{0.000}{PAS}\end{turn}}
	        &AP$\uparrow$	&$\mathsf{0.485_{0.008}}$	&$\mathsf{0.496_{0.004}}$	&$\mathsf{0.508_{0.005}}$	&$\mathsf{0.471_{0.008}}$	&$\mathsf{0.504_{0.005}}$	&$\mathsf{0.437_{0.018}}$	&$\mathsf{0.488_{0.003}}$	&$\mathsf{0.505_{0.012}}$	&$\mathsf{0.555_{0.010}}$\\
	        &1-HL$\uparrow$	&$\mathsf{0.908_{0.002}}$	&$\mathsf{0.927_{0.000}}$	&$\mathsf{0.931_{0.001}}$	&$\mathsf{0.928_{0.001}}$	&$\mathsf{0.93_{0.000}}$	&$\mathsf{0.882_{0.004}}$	&$\mathsf{0.928_{0.001}}$	&$\mathsf{0.929_{0.001}}$	&$\mathsf{0.933_{0.001}}$\\
	        &1-RL$\uparrow$	&$\mathsf{0.745_{0.009}}$	&$\mathsf{0.767_{0.004}}$	&$\mathsf{0.812_{0.004}}$	&$\mathsf{0.761_{0.005}}$	&$\mathsf{0.806_{0.003}}$	&$\mathsf{0.736_{0.015}}$	&$\mathsf{0.783_{0.001}}$	&$\mathsf{0.783_{0.008}}$	&$\mathsf{0.830_{0.005}}$\\
	        &AUC$\uparrow$	&$\mathsf{0.765_{0.01}}$	&$\mathsf{0.786_{0.003}}$	&$\mathsf{0.838_{0.003}}$	&$\mathsf{0.779_{0.004}}$	&$\mathsf{0.832_{0.002}}$	&$\mathsf{0.767_{0.015}}$	&$\mathsf{0.811_{0.001}}$	&$\mathsf{0.809_{0.006}}$	&$\mathsf{0.851_{0.006}}$\\
	        &OE$\downarrow$	&$\mathsf{0.562_{0.008}}$	&$\mathsf{0.557_{0.005}}$	&$\mathsf{0.581_{0.008}}$	&$\mathsf{0.58_{0.011}}$	&$\mathsf{0.581_{0.008}}$	&$\mathsf{0.638_{0.023}}$	&$\mathsf{0.579_{0.006}}$	&$\mathsf{0.573_{0.015}}$	&$\mathsf{0.537_{0.016}}$\\
	        &Cov$\downarrow$	&$\mathsf{0.32_{0.01}}$	&$\mathsf{0.297_{0.004}}$	&$\mathsf{0.241_{0.003}}$	&$\mathsf{0.308_{0.004}}$	&$\mathsf{0.249_{0.003}}$	&$\mathsf{0.323_{0.015}}$	&$\mathsf{0.273_{0.002}}$	&$\mathsf{0.269_{0.006}}$	&$\mathsf{0.217_{0.006}}$\\
	        \rowcolor{LGray}&Ave.R$\downarrow$	&7.17 	&5.33 	&2.83 	&6.67 	&3.83 	&8.83 	&4.83 	&4.00 	&1.00 \\
    \midrule
    \multirow{6}[3]{*}{\begin{turn}{0.000}{ESP}\end{turn}}
	        &AP$\uparrow$	&$\mathsf{0.202_{0.006}}$	&$\mathsf{0.221_{0.002}}$	&$\mathsf{0.289_{0.003}}$	&$\mathsf{0.212_{0.002}}$	&$\mathsf{0.285_{0.003}}$	&$\mathsf{0.244_{0.005}}$	&$\mathsf{0.246_{0.002}}$	&$\mathsf{0.297_{0.002}}$	&$\mathsf{0.314_{0.004}}$\\
	        &1-HL$\uparrow$	&$\mathsf{0.971_{0.002}}$	&$\mathsf{0.982_{0.000}}$	&$\mathsf{0.983_{0.000}}$	&$\mathsf{0.982_{0.000}}$	&$\mathsf{0.983_{0.000}}$	&$\mathsf{0.972_{0.000}}$	&$\mathsf{0.983_{0.000}}$	&$\mathsf{0.983_{0.000}}$	&$\mathsf{0.983_{0.000}}$\\
	        &1-RL$\uparrow$	&$\mathsf{0.772_{0.006}}$	&$\mathsf{0.78_{0.004}}$	&$\mathsf{0.832_{0.001}}$	&$\mathsf{0.781_{0.001}}$	&$\mathsf{0.829_{0.001}}$	&$\mathsf{0.808_{0.002}}$	&$\mathsf{0.818_{0.002}}$	&$\mathsf{0.832_{0.001}}$	&$\mathsf{0.848_{0.002}}$\\
	        &AUC$\uparrow$	&$\mathsf{0.777_{0.006}}$	&$\mathsf{0.784_{0.004}}$	&$\mathsf{0.836_{0.001}}$	&$\mathsf{0.785_{0.001}}$	&$\mathsf{0.833_{0.002}}$	&$\mathsf{0.813_{0.002}}$	&$\mathsf{0.824_{0.002}}$	&$\mathsf{0.836_{0.001}}$	&$\mathsf{0.853_{0.002}}$\\
	        &OE$\downarrow$	&$\mathsf{0.738_{0.018}}$	&$\mathsf{0.683_{0.005}}$	&$\mathsf{0.604_{0.005}}$	&$\mathsf{0.706_{0.006}}$	&$\mathsf{0.611_{0.004}}$	&$\mathsf{0.657_{0.013}}$	&$\mathsf{0.661_{0.003}}$	&$\mathsf{0.561_{0.007}}$	&$\mathsf{0.538_{0.015}}$\\
	        &Cov$\downarrow$	&$\mathsf{0.503_{0.011}}$	&$\mathsf{0.504_{0.006}}$	&$\mathsf{0.426_{0.004}}$	&$\mathsf{0.512_{0.003}}$	&$\mathsf{0.433_{0.005}}$	&$\mathsf{0.452_{0.004}}$	&$\mathsf{0.429_{0.003}}$	&$\mathsf{0.407_{0.003}}$	&$\mathsf{0.370_{0.005}}$\\
	        \rowcolor{LGray}&Ave.R$\downarrow$	&8.67 	&7.33 	&2.33 	&7.50 	&3.67 	&6.17 	&4.33 	&1.83 	&1.00 \\
    \midrule
    \multirow{6}[3]{*}{\begin{turn}{0.000}{IAP}\end{turn}}
	        &AP$\uparrow$	&$\mathsf{0.224_{0.007}}$	&$\mathsf{0.256_{0.002}}$	&$\mathsf{0.305_{0.004}}$	&$\mathsf{0.234_{0.003}}$	&$\mathsf{0.304_{0.004}}$	&$\mathsf{0.237_{0.003}}$	&$\mathsf{0.261_{0.001}}$	&$\mathsf{0.323_{0.001}}$	&$\mathsf{0.346_{0.004}}$\\
	        &1-HL$\uparrow$	&$\mathsf{0.965_{0.002}}$	&$\mathsf{0.98_{0.000}}$	&$\mathsf{0.981_{0.000}}$	&$\mathsf{0.98_{0.000}}$	&$\mathsf{0.981_{0.000}}$	&$\mathsf{0.969_{0.000}}$	&$\mathsf{0.98_{0.000}}$	&$\mathsf{0.981_{0.000}}$	&$\mathsf{0.981_{0.000}}$\\
	        &1-RL$\uparrow$	&$\mathsf{0.806_{0.005}}$	&$\mathsf{0.825_{0.002}}$	&$\mathsf{0.862_{0.002}}$	&$\mathsf{0.823_{0.002}}$	&$\mathsf{0.861_{0.002}}$	&$\mathsf{0.833_{0.002}}$	&$\mathsf{0.848_{0.001}}$	&$\mathsf{0.873_{0.001}}$	&$\mathsf{0.888_{0.002}}$\\
	        &AUC$\uparrow$	&$\mathsf{0.807_{0.005}}$	&$\mathsf{0.83_{0.001}}$	&$\mathsf{0.864_{0.002}}$	&$\mathsf{0.825_{0.001}}$	&$\mathsf{0.863_{0.001}}$	&$\mathsf{0.835_{0.001}}$	&$\mathsf{0.85_{0.001}}$	&$\mathsf{0.874_{0.000}}$	&$\mathsf{0.889_{0.002}}$\\
	        &OE$\downarrow$	&$\mathsf{0.7_{0.031}}$	&$\mathsf{0.622_{0.007}}$	&$\mathsf{0.568_{0.008}}$	&$\mathsf{0.66_{0.006}}$	&$\mathsf{0.571_{0.009}}$	&$\mathsf{0.648_{0.008}}$	&$\mathsf{0.61_{0.005}}$	&$\mathsf{0.532_{0.002}}$	&$\mathsf{0.515_{0.006}}$\\
	        &Cov$\downarrow$	&$\mathsf{0.477_{0.009}}$	&$\mathsf{0.466_{0.003}}$	&$\mathsf{0.403_{0.004}}$	&$\mathsf{0.471_{0.004}}$	&$\mathsf{0.403_{0.004}}$	&$\mathsf{0.436_{0.005}}$	&$\mathsf{0.408_{0.004}}$	&$\mathsf{0.351_{0.001}}$	&$\mathsf{0.315_{0.005}}$\\
	        \rowcolor{LGray}&Ave.R$\downarrow$	&9.00 	&6.33 	&2.67 	&7.50 	&3.33 	&6.67 	&5.00 	&1.83 	&1.00 \\
    \midrule
    \multirow{6}[3]{*}{\begin{turn}{0.000}{MIR}\end{turn}}
	        &AP$\uparrow$	&$\mathsf{0.505_{0.008}}$	&$\mathsf{0.537_{0.002}}$	&$\mathsf{0.57_{0.002}}$	&$\mathsf{0.514_{0.006}}$	&$\mathsf{0.553_{0.002}}$	&$\mathsf{0.49_{0.012}}$	&$\mathsf{0.551_{0.002}}$	&$\mathsf{0.589_{0.005}}$	&$\mathsf{0.602_{0.007}}$\\
	        &1-HL$\uparrow$	&$\mathsf{0.853_{0.004}}$	&$\mathsf{0.874_{0.001}}$	&$\mathsf{0.886_{0.001}}$	&$\mathsf{0.878_{0.001}}$	&$\mathsf{0.885_{0.001}}$	&$\mathsf{0.839_{0.002}}$	&$\mathsf{0.882_{0.001}}$	&$\mathsf{0.888_{0.002}}$	&$\mathsf{0.890_{0.001}}$\\
	        &1-RL$\uparrow$	&$\mathsf{0.821_{0.003}}$	&$\mathsf{0.832_{0.001}}$	&$\mathsf{0.856_{0.001}}$	&$\mathsf{0.831_{0.003}}$	&$\mathsf{0.856_{0.001}}$	&$\mathsf{0.803_{0.008}}$	&$\mathsf{0.844_{0.001}}$	&$\mathsf{0.863_{0.004}}$	&$\mathsf{0.875_{0.003}}$\\
	        &AUC$\uparrow$	&$\mathsf{0.81_{0.004}}$	&$\mathsf{0.828_{0.001}}$	&$\mathsf{0.846_{0.001}}$	&$\mathsf{0.828_{0.003}}$	&$\mathsf{0.844_{0.001}}$	&$\mathsf{0.787_{0.012}}$	&$\mathsf{0.837_{0.001}}$	&$\mathsf{0.849_{0.004}}$	&$\mathsf{0.860_{0.003}}$\\
	        &OE$\downarrow$	&$\mathsf{0.495_{0.02}}$	&$\mathsf{0.448_{0.005}}$	&$\mathsf{0.369_{0.004}}$	&$\mathsf{0.49_{0.008}}$	&$\mathsf{0.393_{0.004}}$	&$\mathsf{0.489_{0.022}}$	&$\mathsf{0.415_{0.003}}$	&$\mathsf{0.363_{0.007}}$	&$\mathsf{0.347_{0.009}}$\\
	        &Cov$\downarrow$	&$\mathsf{0.41_{0.005}}$	&$\mathsf{0.395_{0.003}}$	&$\mathsf{0.36_{0.001}}$	&$\mathsf{0.396_{0.005}}$	&$\mathsf{0.364_{0.001}}$	&$\mathsf{0.428_{0.013}}$	&$\mathsf{0.369_{0.002}}$	&$\mathsf{0.348_{0.007}}$	&$\mathsf{0.330_{0.004}}$\\
	        \rowcolor{LGray}&Ave.R$\downarrow$	&8.17 	&6.17 	&3.00 	&6.83 	&3.83 	&8.67 	&5.00 	&2.00 	&1.00 \\
\bottomrule
\end{tabular}
\vspace{-0.25cm}
\end{table*}

\subsubsection{Competitors}
To evaluate the performance of our method, we select the 8 most relevant methods, \textbf{C2AE} \cite{yeh2017learning}, \textbf{GLOCAL} \cite{zhu2017multi}, \textbf{CDMM} \cite{zhao2021consistency}, \textbf{DM2L} \cite{ma2021expand}, \textbf{LVSL} \cite{zhao2022non}, \textbf{NAIM3L} \cite{li2022concise}, \textbf{iMVWL} \cite{tan2018incomplete}, and \textbf{DICNet} \cite{liu2023dicnet}, to compare with our RANK on the five incomplete multi-view missing multi-label datasets. Most of the above methods are introduced in Section \ref{sec2}, except the C2AE, which is a representative deep neural networks based multi-label classification model with two networks for exploiting feature and label embedding representation. And the C2AE can only handle the single-view multi-label classification task with missing labels. For such single-view multi-label methods, C2AE, GLOCAL, and DM2L, we execute the algorithms on each view separately and report the optimal result. Simple statistics of comparison methods are shown in Table \ref{tab:methods}.

Of these 8 methods, only iMVWL \cite{tan2018incomplete}, NAIM3L \cite{li2022concise}, and DICNet \cite{liu2023dicnet} are compatible with the dual incompleteness of views and labels, therefore, following works \cite{tan2018incomplete,li2022concise,liu2023dicnet}, we make necessary modifications to them to meet the experimental settings. Specifically, for methods that cannot handle missing views (CDMM, C2AE, GLOCAL, DM2L, and LVSL), we simply fill the missing views with the average feature of the available views; for methods that cannot cope with missing labels (CDMM and LSVL), we set the missing labels as zero.

\subsubsection{Evaluation Metrics}
In keeping with existing iM3C works \cite{tan2018incomplete,li2022concise,liu2023dicnet}, we choose 4 popular multi-label classification metrics, i.e., Ranking Loss (\textbf{RL}), Average Precision (\textbf{AP}), Hamming Loss (\textbf{HL}), and adapted area under curve (\textbf{AUC}) to evaluate these methods. Note that, like above methods, we report '\textbf{1-RL}' and '\textbf{1-HL}' instead of RL and HL due to the relatively small values of them. In addition, we introduce the OneError (\textbf{OE}) and Coverage (\textbf{Cov}) metrics commonly used in multi-label classification methods in the experiments \cite{zhang2013review,chen2019multi}. For the first 4 metrics, the higher the value the better the performance. But for OE and Cov, a lower value is what we would expect. Finally, to intuitively compare the methods, we compute the average ranking of each method on all six metrics.
\subsubsection{Implementation Details}
As a deep learning model, our RANK is coded by popular Python and Pytorch framework. We set learning rate as $0.1$ and select the Stochastic Gradient Descent (SGD) optimizer to train our model. The batch size and momentum are set to 128 and 0.9 for all datasets. Our RANK \footnote{Code: \url{https://github.com/justsmart/RANK}} runs on Ubuntu system with an RTX 3090 gpu and an i7-12900k cpu. Besides, to avoid the randomness in partitioning and constructing incomplete datasets, we repeat above random operations many times to report the mean and standard deviation of results.	

\begin{figure*}[t!]
		\centering
		\subfloat[Corel5k]{
			\label{fig:view-mis-cor}
			\includegraphics[width=0.24\textwidth]{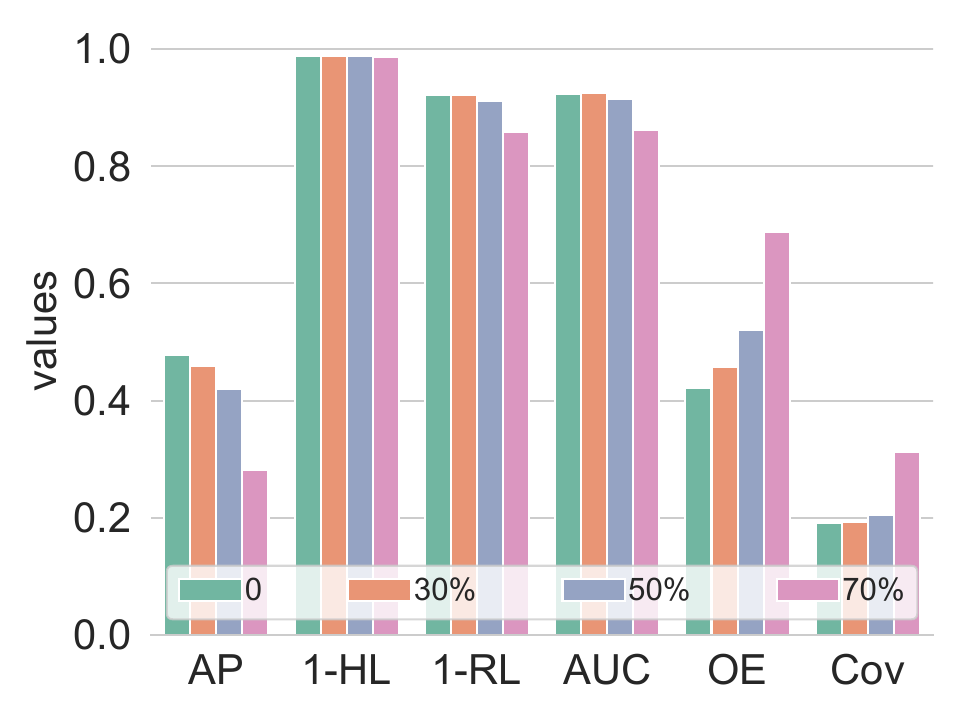}
		}
		\subfloat[Pascal07]{
			\label{fig:view-mis-pas}
			\includegraphics[width=0.24\textwidth]{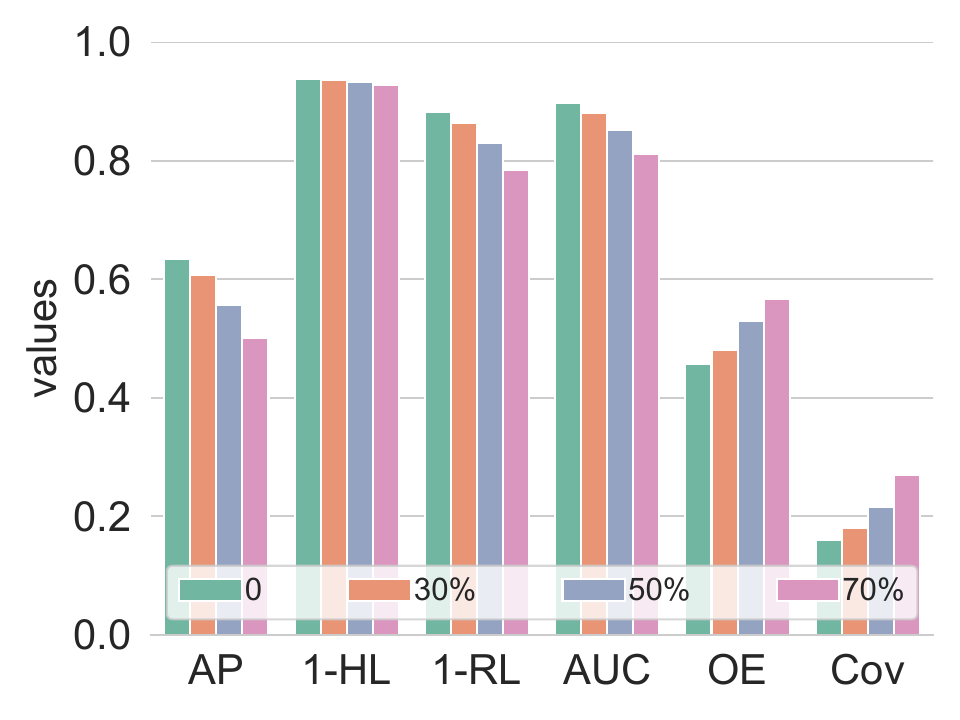}
		}
		\subfloat[Corel5k]{
			\label{fig:label-mis-cor}
			\includegraphics[width=0.24\textwidth]{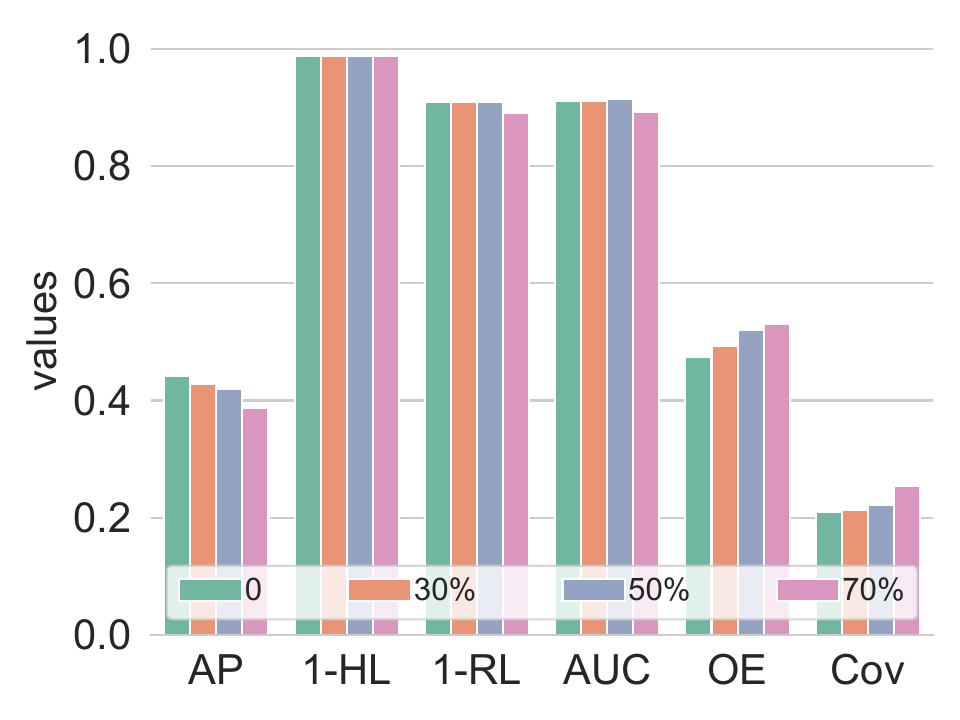}
		}
		\subfloat[Pascal07]{
			\label{fig:label-mis-pas}
			\includegraphics[width=0.24\textwidth]{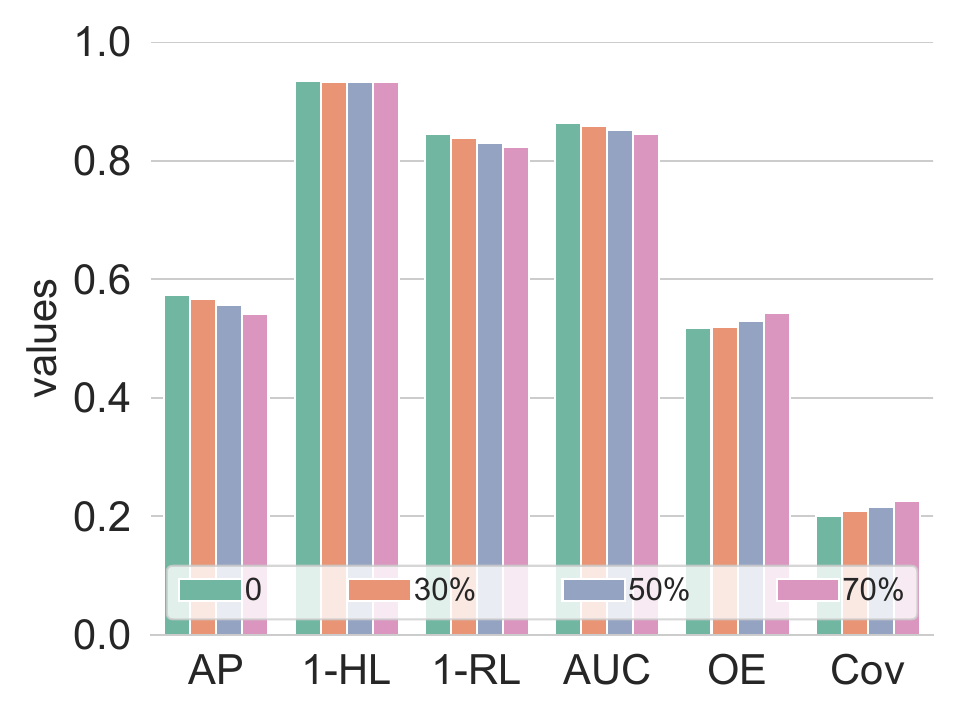}
		}
		
		\caption{Performance trends under different missing rates on Corel5k and Pascal07 with (ab) different missing-view ratios and 50\% missing labels; (cd) 50\% missing views and different missing-label ratios.}
		\label{fig:rates}
		
\end{figure*}

\begin{figure*}[!t]
\centering
\subfloat[30\% view-missing ratio and 30\% label-missing ratio]{
\label{fig:alla}
\includegraphics[width=0.24\textwidth]{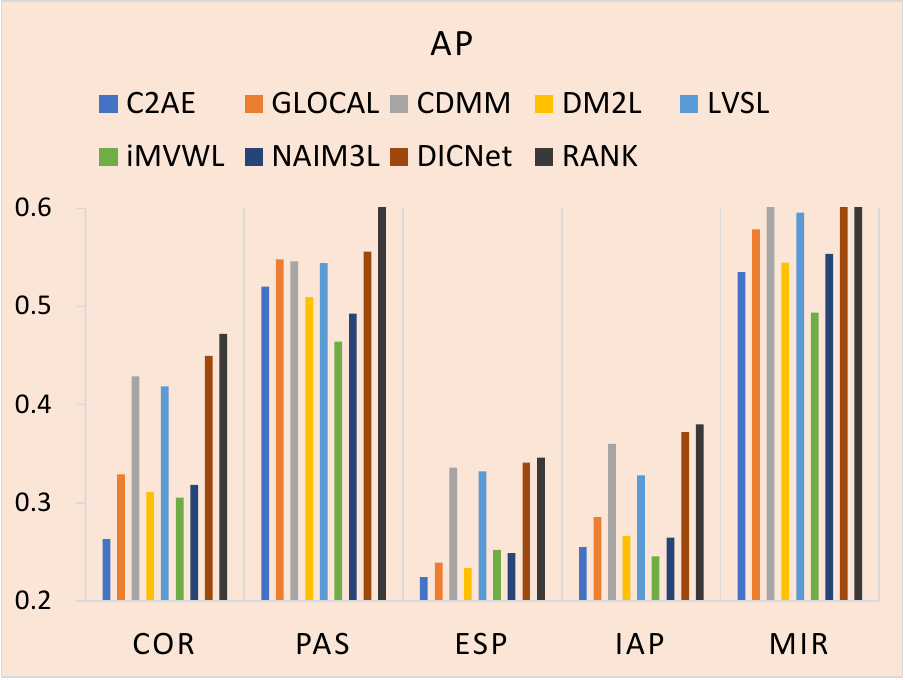}
\includegraphics[width=0.24\textwidth]{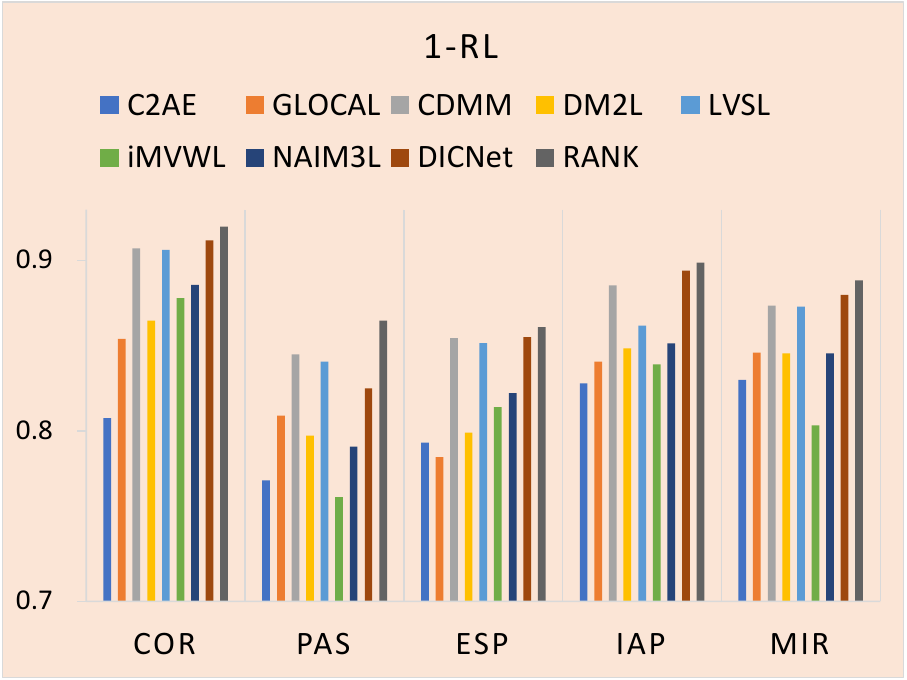}
\includegraphics[width=0.24\textwidth]{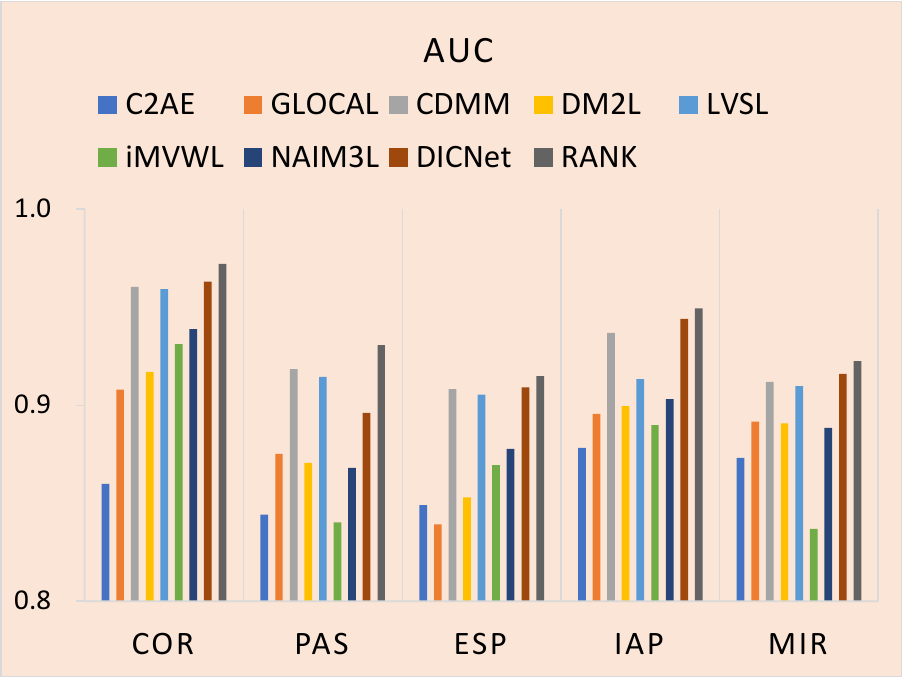}
\includegraphics[width=0.24\textwidth]{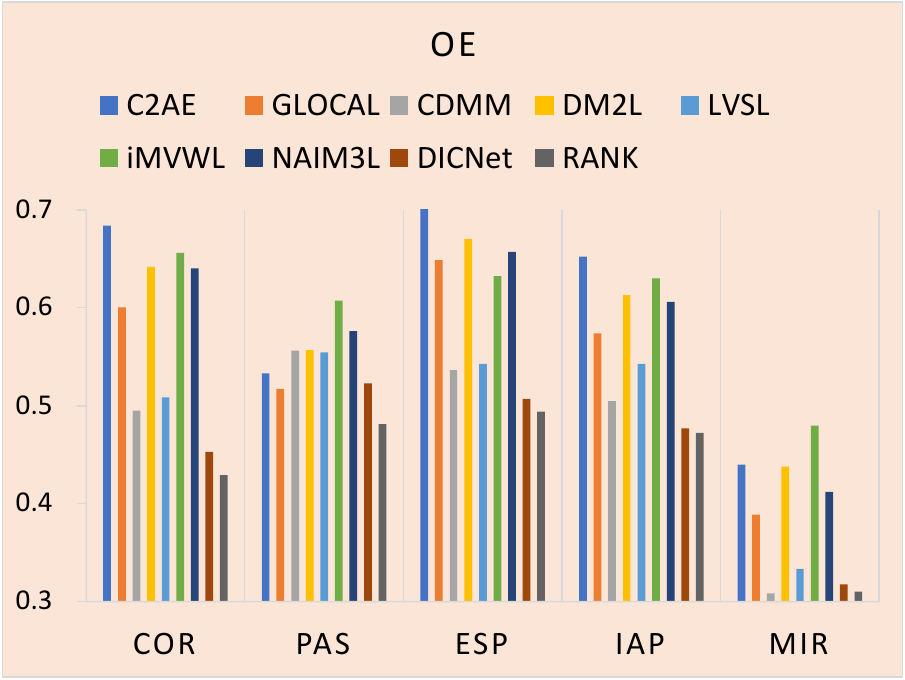}
}
\quad
\subfloat[full views and 50\% label-missing ratio]{
\label{fig:allb}
\includegraphics[width=0.24\textwidth]{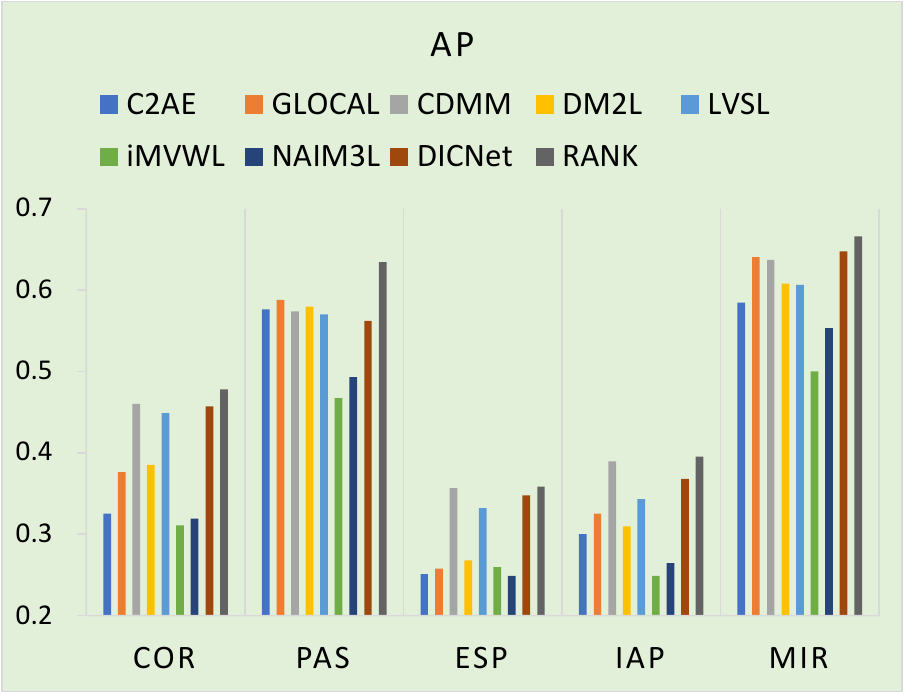}
\includegraphics[width=0.24\textwidth]{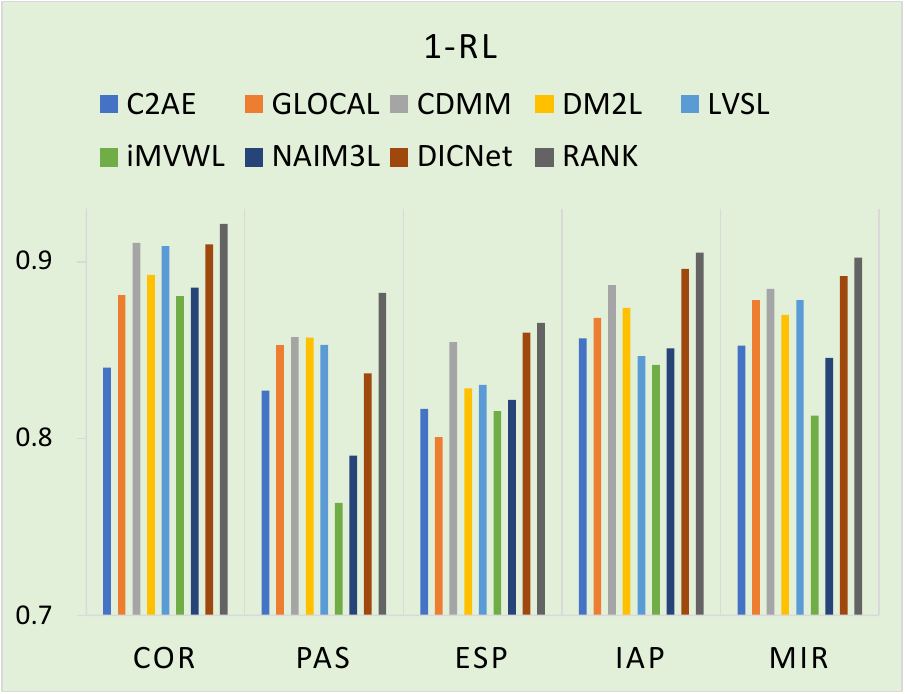}
\includegraphics[width=0.24\textwidth]{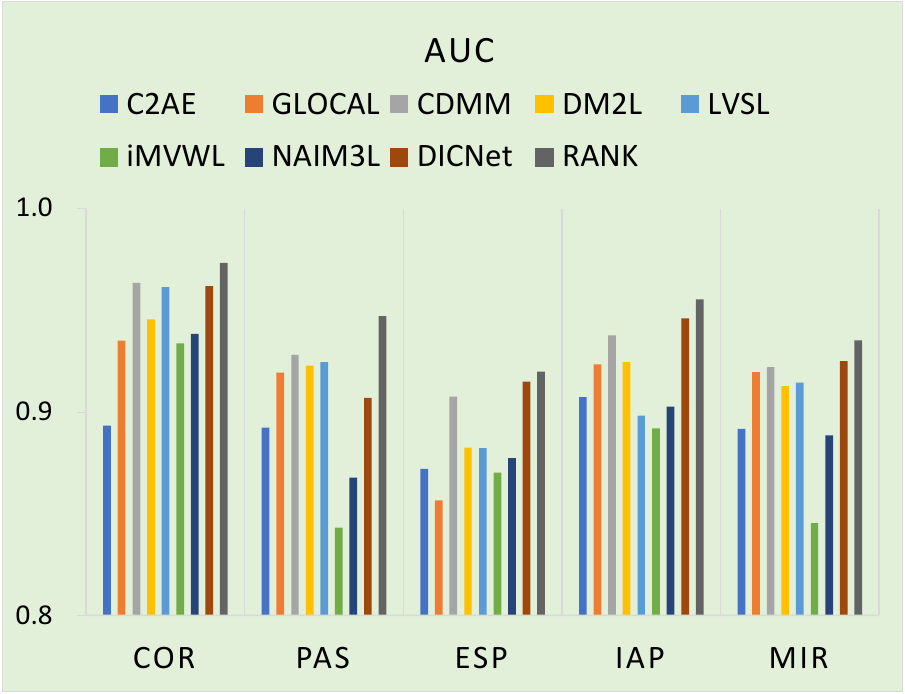}
\includegraphics[width=0.24\textwidth]{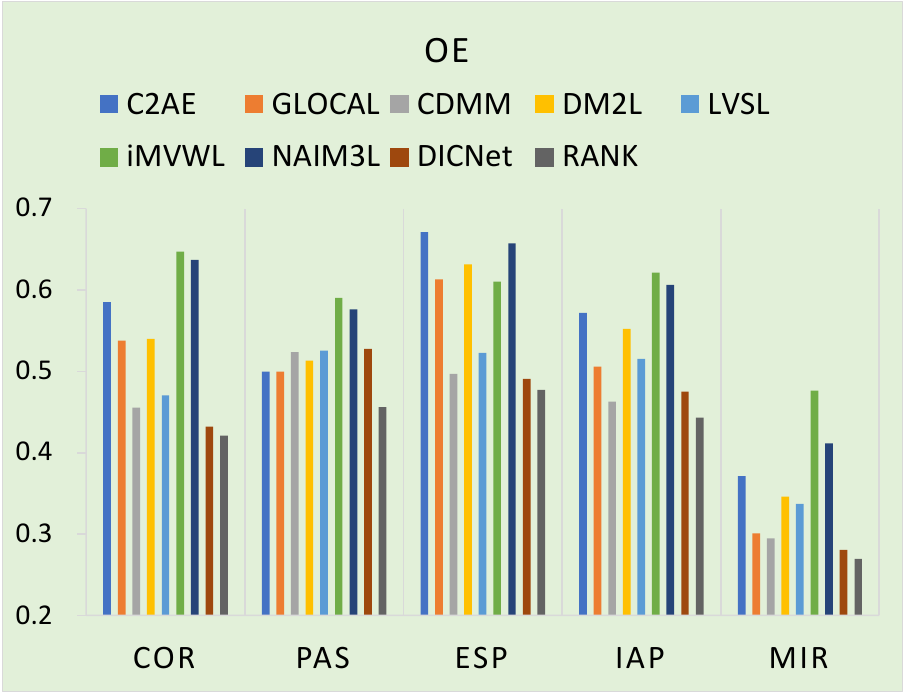}
}
\quad
\subfloat[50\% view-missing ratio and full labels]{
\label{fig:allc}
\includegraphics[width=0.24\textwidth]{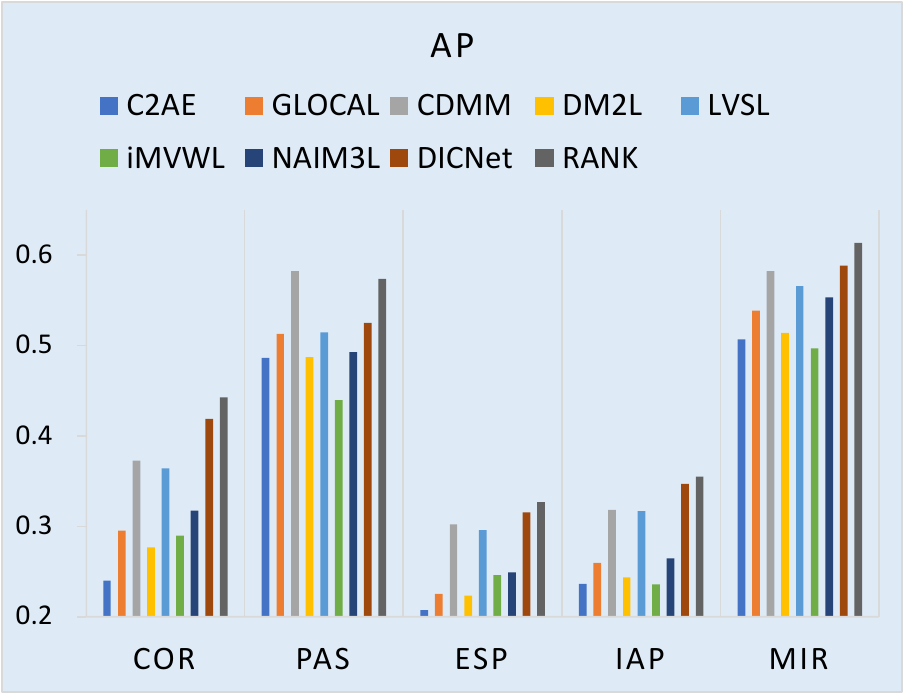}
\includegraphics[width=0.24\textwidth]{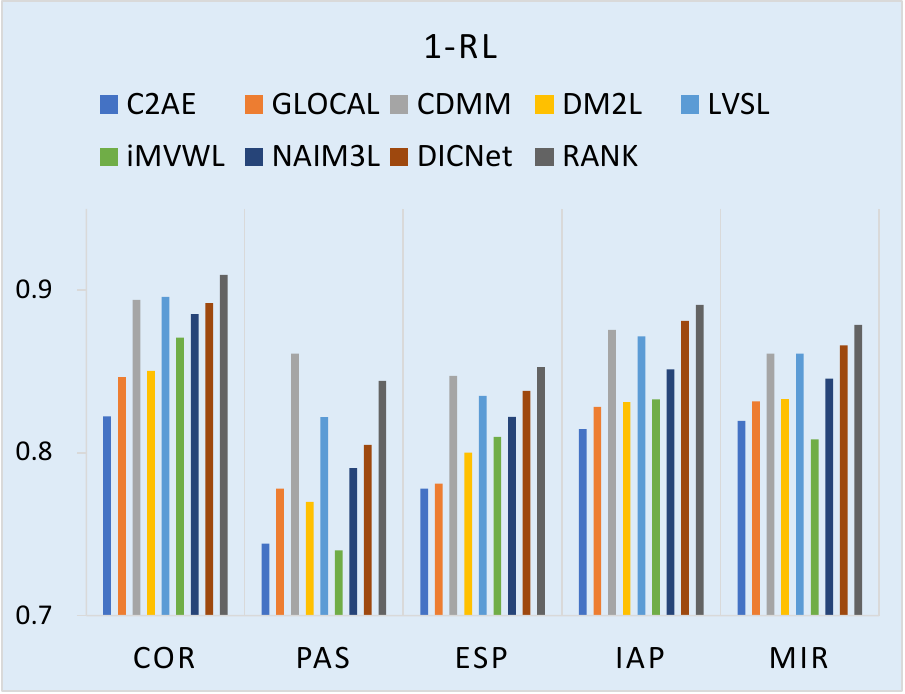}
\includegraphics[width=0.24\textwidth]{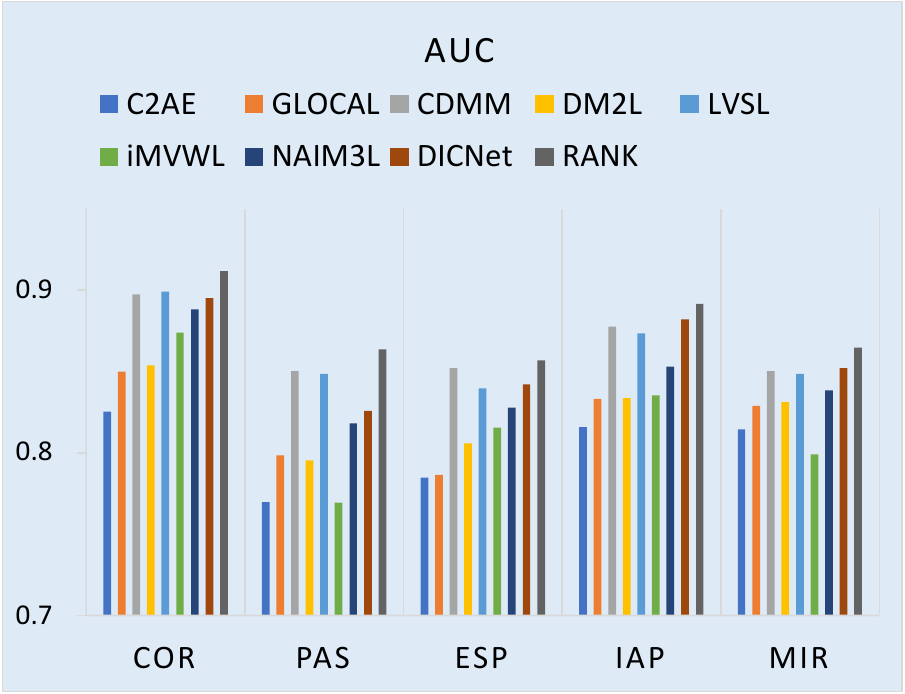}
\includegraphics[width=0.24\textwidth]{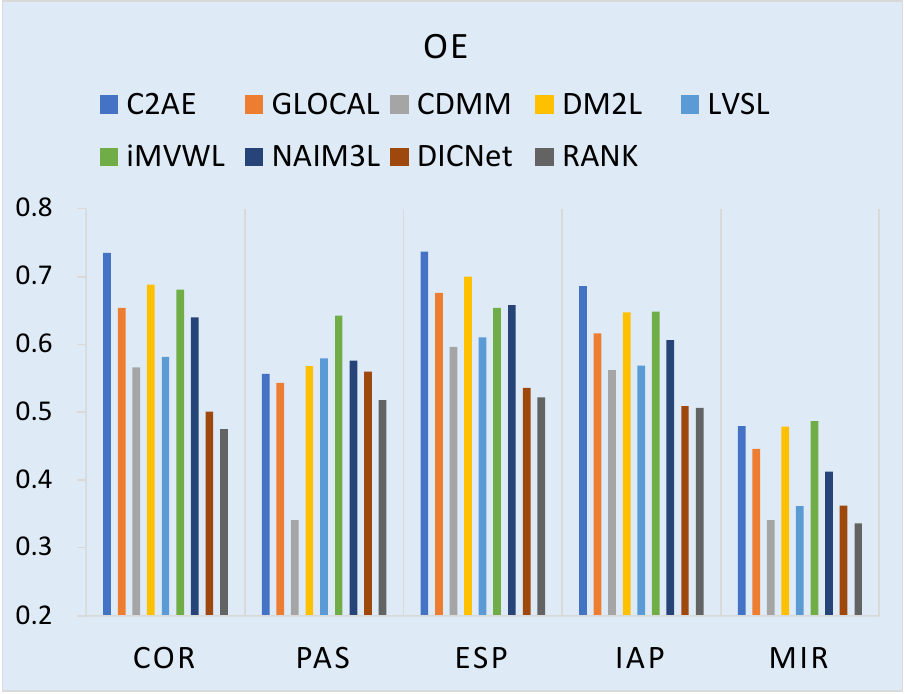}
}
\caption{More experimental results on different missing ratios.}
\label{fig:all}
\vspace{-0.3cm}
\end{figure*}

\begin{table*}[t!]
\renewcommand{\arraystretch}{1.1}
\centering
\sc
\caption{Experimental results of nine methods on the five datasets with full views and labels. $\uparrow$ indicates that the higher the value, the better the performance, and $\downarrow$ is the opposite. 'Ave.R' represents the average ranking of the corresponding method on all six metrics.}
\label{table:com}
\small
    \begin{tabular}{p{0.7cm}p{0.95cm}cccccccc>{\columncolor{Gray}}c}
	\toprule
    Data & Metric & C2AE   & GLOCAL    & CDMM  &DM2L &LVSL &iMVWL&NAIM3L&DICNet&RANK\\
    \midrule	
    \multirow{6}[3]{*}{\begin{turn}{0}{COR}\end{turn}}
              &AP$\uparrow$	&$\mathsf{0.353_{0.033}}$	&$\mathsf{0.386_{0.006}}$	&$\mathsf{0.489_{0.004}}$	&$\mathsf{0.401_{0.006}}$	&$\mathsf{0.482_{0.003}}$	&$\mathsf{0.313_{0.005}}$	&$\mathsf{0.327_{0.004}}$	&$\mathsf{0.509_{0.002}}$	&$\mathsf{0.524_{0.008}}$\\
              &1-HL$\uparrow$	&$\mathsf{0.969_{0.002}}$	&$\mathsf{0.987_{0.000}}$	&$\mathsf{0.988_{0.000}}$	&$\mathsf{0.987_{0.000}}$	&$\mathsf{0.988_{0.000}}$	&$\mathsf{0.979_{0.000}}$	&$\mathsf{0.987_{0.000}}$	&$\mathsf{0.989_{0.000}}$	&$\mathsf{0.989_{0.000}}$\\
              &1-RL$\uparrow$	&$\mathsf{0.870_{0.020}}$	&$\mathsf{0.891_{0.002}}$	&$\mathsf{0.926_{0.001}}$	&$\mathsf{0.905_{0.002}}$	&$\mathsf{0.927_{0.001}}$	&$\mathsf{0.884_{0.002}}$	&$\mathsf{0.890_{0.002}}$	&$\mathsf{0.929_{0.001}}$	&$\mathsf{0.932_{0.003}}$\\
              &AUC$\uparrow$	&$\mathsf{0.873_{0.020}}$	&$\mathsf{0.895_{0.002}}$	&$\mathsf{0.929_{0.001}}$	&$\mathsf{0.908_{0.002}}$	&$\mathsf{0.930_{0.001}}$	&$\mathsf{0.887_{0.002}}$	&$\mathsf{0.893_{0.002}}$	&$\mathsf{0.931_{0.000}}$	&$\mathsf{0.934_{0.003}}$\\
              &OE$\downarrow$	&$\mathsf{0.566_{0.042}}$	&$\mathsf{0.523_{0.012}}$	&$\mathsf{0.433_{0.007}}$	&$\mathsf{0.531_{0.012}}$	&$\mathsf{0.446_{0.003}}$	&$\mathsf{0.644_{0.010}}$	&$\mathsf{0.629_{0.009}}$	&$\mathsf{0.385_{0.007}}$	&$\mathsf{0.373_{0.013}}$\\
              &Cov$\downarrow$	&$\mathsf{0.274_{0.034}}$	&$\mathsf{0.250_{0.004}}$	&$\mathsf{0.188_{0.003}}$	&$\mathsf{0.228_{0.005}}$	&$\mathsf{0.181_{0.003}}$	&$\mathsf{0.261_{0.005}}$	&$\mathsf{0.249_{0.005}}$	&$\mathsf{0.178_{0.001}}$	&$\mathsf{0.166_{0.006}}$\\
              \rowcolor{LGray}&Ave.R$\downarrow$	&8.33 	&5.83 	&3.50 	&5.17 	&3.33 	&8.33 	&6.83 	&1.83 	&1.00 \\
    \midrule
    \multirow{6}[3]{*}{\begin{turn}{0.000}{PAS}\end{turn}}
	        &AP$\uparrow$	&$\mathsf{0.584_{0.007}}$	&$\mathsf{0.610_{0.005}}$	&$\mathsf{0.589_{0.006}}$	&$\mathsf{0.578_{0.009}}$	&$\mathsf{0.588_{0.006}}$	&$\mathsf{0.468_{0.004}}$	&$\mathsf{0.496_{0.003}}$	&$\mathsf{0.608_{0.003}}$	&$\mathsf{0.656_{0.007}}$\\
	        &1-HL$\uparrow$	&$\mathsf{0.921_{0.002}}$	&$\mathsf{0.926_{0.000}}$	&$\mathsf{0.931_{0.001}}$	&$\mathsf{0.932_{0.001}}$	&$\mathsf{0.934_{0.001}}$	&$\mathsf{0.888_{0.001}}$	&$\mathsf{0.929_{0.001}}$	&$\mathsf{0.937_{0.001}}$	&$\mathsf{0.940_{0.001}}$\\
	        &1-RL$\uparrow$	&$\mathsf{0.835_{0.006}}$	&$\mathsf{0.866_{0.002}}$	&$\mathsf{0.873_{0.003}}$	&$\mathsf{0.856_{0.003}}$	&$\mathsf{0.871_{0.003}}$	&$\mathsf{0.763_{0.007}}$	&$\mathsf{0.795_{0.001}}$	&$\mathsf{0.859_{0.003}}$	&$\mathsf{0.891_{0.004}}$\\
	        &AUC$\uparrow$	&$\mathsf{0.851_{0.006}}$	&$\mathsf{0.879_{0.002}}$	&$\mathsf{0.893_{0.002}}$	&$\mathsf{0.874_{0.003}}$	&$\mathsf{0.891_{0.002}}$	&$\mathsf{0.793_{0.005}}$	&$\mathsf{0.822_{0.001}}$	&$\mathsf{0.876_{0.002}}$	&$\mathsf{0.902_{0.005}}$\\
	        &OE$\downarrow$	&$\mathsf{0.490_{0.009}}$	&$\mathsf{0.478_{0.008}}$	&$\mathsf{0.518_{0.008}}$	&$\mathsf{0.511_{0.010}}$	&$\mathsf{0.517_{0.008}}$	&$\mathsf{0.588_{0.005}}$	&$\mathsf{0.574_{0.006}}$	&$\mathsf{0.474_{0.005}}$	&$\mathsf{0.426_{0.013}}$\\
	        &Cov$\downarrow$	&$\mathsf{0.217_{0.007}}$	&$\mathsf{0.184_{0.003}}$	&$\mathsf{0.172_{0.003}}$	&$\mathsf{0.193_{0.004}}$	&$\mathsf{0.175_{0.003}}$	&$\mathsf{0.296_{0.007}}$	&$\mathsf{0.259_{0.002}}$	&$\mathsf{0.185_{0.004}}$	&$\mathsf{0.152_{0.005}}$\\
	        \rowcolor{LGray}&Ave.R$\downarrow$	&6.50 	&4.00 	&3.67 	&5.67 	&3.83 	&9.00 	&7.67 	&3.67 	&1.00 \\
    \midrule
    \multirow{6}[3]{*}{\begin{turn}{0.000}{ESP}\end{turn}}
	        &AP$\uparrow$	&$\mathsf{0.269_{0.003}}$	&$\mathsf{0.264_{0.002}}$	&$\mathsf{0.392_{0.002}}$	&$\mathsf{0.268_{0.002}}$	&$\mathsf{0.363_{0.002}}$	&$\mathsf{0.260_{0.004}}$	&$\mathsf{0.251_{0.002}}$	&$\mathsf{0.391_{0.001}}$	&$\mathsf{0.388_{0.003}}$\\
	        &1-HL$\uparrow$	&$\mathsf{0.958_{0.002}}$	&$\mathsf{0.983_{0.000}}$	&$\mathsf{0.983_{0.000}}$	&$\mathsf{0.983_{0.000}}$	&$\mathsf{0.983_{0.000}}$	&$\mathsf{0.972_{0.000}}$	&$\mathsf{0.983_{0.000}}$	&$\mathsf{0.984_{0.000}}$	&$\mathsf{0.984_{0.000}}$\\
	        &1-RL$\uparrow$	&$\mathsf{0.832_{0.002}}$	&$\mathsf{0.804_{0.003}}$	&$\mathsf{0.875_{0.001}}$	&$\mathsf{0.841_{0.001}}$	&$\mathsf{0.836_{0.001}}$	&$\mathsf{0.817_{0.001}}$	&$\mathsf{0.825_{0.001}}$	&$\mathsf{0.871_{0.003}}$	&$\mathsf{0.877_{0.001}}$\\
	        &AUC$\uparrow$	&$\mathsf{0.837_{0.002}}$	&$\mathsf{0.810_{0.003}}$	&$\mathsf{0.878_{0.001}}$	&$\mathsf{0.846_{0.001}}$	&$\mathsf{0.838_{0.001}}$	&$\mathsf{0.822_{0.001}}$	&$\mathsf{0.830_{0.002}}$	&$\mathsf{0.874_{0.004}}$	&$\mathsf{0.881_{0.002}}$\\
	        &OE$\downarrow$	&$\mathsf{0.646_{0.006}}$	&$\mathsf{0.602_{0.004}}$	&$\mathsf{0.468_{0.004}}$	&$\mathsf{0.651_{0.004}}$	&$\mathsf{0.495_{0.004}}$	&$\mathsf{0.610_{0.011}}$	&$\mathsf{0.655_{0.003}}$	&$\mathsf{0.451_{0.002}}$	&$\mathsf{0.454_{0.006}}$\\
	        &Cov$\downarrow$	&$\mathsf{0.386_{0.005}}$	&$\mathsf{0.456_{0.005}}$	&$\mathsf{0.336_{0.003}}$	&$\mathsf{0.389_{0.002}}$	&$\mathsf{0.423_{0.003}}$	&$\mathsf{0.434_{0.003}}$	&$\mathsf{0.413_{0.003}}$	&$\mathsf{0.326_{0.006}}$	&$\mathsf{0.312_{0.004}}$\\
	        \rowcolor{LGray}&Ave.R$\downarrow$	&6.17 	&7.00 	&2.33 	&5.00 	&4.67 	&7.67 	&6.83 	&2.00 	&1.50 \\
    \midrule
    \multirow{6}[3]{*}{\begin{turn}{0.000}{IAP}\end{turn}}
	        &AP$\uparrow$	&$\mathsf{0.316_{0.004}}$	&$\mathsf{0.330_{0.003}}$	&$\mathsf{0.424_{0.003}}$	&$\mathsf{0.323_{0.008}}$	&$\mathsf{0.383_{0.003}}$	&$\mathsf{0.250_{0.004}}$	&$\mathsf{0.267_{0.001}}$	&$\mathsf{0.418_{0.006}}$	&$\mathsf{0.428_{0.005}}$\\
	        &1-HL$\uparrow$	&$\mathsf{0.953_{0.001}}$	&$\mathsf{0.980_{0.000}}$	&$\mathsf{0.982_{0.000}}$	&$\mathsf{0.981_{0.000}}$	&$\mathsf{0.982_{0.000}}$	&$\mathsf{0.969_{0.000}}$	&$\mathsf{0.980_{0.000}}$	&$\mathsf{0.982_{0.000}}$	&$\mathsf{0.982_{0.000}}$\\
	        &1-RL$\uparrow$	&$\mathsf{0.868_{0.001}}$	&$\mathsf{0.871_{0.002}}$	&$\mathsf{0.905_{0.001}}$	&$\mathsf{0.879_{0.004}}$	&$\mathsf{0.872_{0.001}}$	&$\mathsf{0.842_{0.002}}$	&$\mathsf{0.854_{0.001}}$	&$\mathsf{0.912_{0.003}}$	&$\mathsf{0.914_{0.002}}$\\
	        &AUC$\uparrow$	&$\mathsf{0.869_{0.002}}$	&$\mathsf{0.877_{0.002}}$	&$\mathsf{0.906_{0.001}}$	&$\mathsf{0.881_{0.004}}$	&$\mathsf{0.873_{0.001}}$	&$\mathsf{0.843_{0.002}}$	&$\mathsf{0.855_{0.001}}$	&$\mathsf{0.911_{0.003}}$	&$\mathsf{0.915_{0.003}}$\\
	        &OE$\downarrow$	&$\mathsf{0.559_{0.010}}$	&$\mathsf{0.501_{0.008}}$	&$\mathsf{0.441_{0.007}}$	&$\mathsf{0.539_{0.008}}$	&$\mathsf{0.495_{0.006}}$	&$\mathsf{0.620_{0.010}}$	&$\mathsf{0.604_{0.004}}$	&$\mathsf{0.419_{0.001}}$	&$\mathsf{0.419_{0.008}}$\\
	        &Cov$\downarrow$	&$\mathsf{0.348_{0.002}}$	&$\mathsf{0.362_{0.005}}$	&$\mathsf{0.300_{0.004}}$	&$\mathsf{0.352_{0.010}}$	&$\mathsf{0.380_{0.003}}$	&$\mathsf{0.419_{0.005}}$	&$\mathsf{0.392_{0.003}}$	&$\mathsf{0.269_{0.009}}$	&$\mathsf{0.255_{0.005}}$\\
	        \rowcolor{LGray}&Ave.R$\downarrow$	&6.83 	&5.50 	&2.50 	&5.00 	&4.50 	&8.83 	&7.67 	&1.83 	&1.00 \\
    \midrule
    \multirow{6}[3]{*}{\begin{turn}{0.000}{MIR}\end{turn}}
	        &AP$\uparrow$	&$\mathsf{0.593_{0.004}}$	&$\mathsf{0.643_{0.003}}$	&$\mathsf{0.665_{0.003}}$	&$\mathsf{0.613_{0.005}}$	&$\mathsf{0.638_{0.002}}$	&$\mathsf{0.493_{0.022}}$	&$\mathsf{0.555_{0.002}}$	&$\mathsf{0.659_{0.004}}$	&$\mathsf{0.680_{0.005}}$\\
	        &1-HL$\uparrow$	&$\mathsf{0.869_{0.002}}$	&$\mathsf{0.874_{0.001}}$	&$\mathsf{0.896_{0.001}}$	&$\mathsf{0.890_{0.001}}$	&$\mathsf{0.895_{0.001}}$	&$\mathsf{0.840_{0.004}}$	&$\mathsf{0.882_{0.001}}$	&$\mathsf{0.902_{0.001}}$	&$\mathsf{0.905_{0.001}}$\\
	        &1-RL$\uparrow$	&$\mathsf{0.854_{0.003}}$	&$\mathsf{0.876_{0.006}}$	&$\mathsf{0.894_{0.001}}$	&$\mathsf{0.870_{0.003}}$	&$\mathsf{0.890_{0.001}}$	&$\mathsf{0.806_{0.015}}$	&$\mathsf{0.847_{0.001}}$	&$\mathsf{0.895_{0.001}}$	&$\mathsf{0.907_{0.002}}$\\
	        &AUC$\uparrow$	&$\mathsf{0.845_{0.003}}$	&$\mathsf{0.869_{0.005}}$	&$\mathsf{0.880_{0.001}}$	&$\mathsf{0.864_{0.003}}$	&$\mathsf{0.875_{0.001}}$	&$\mathsf{0.789_{0.022}}$	&$\mathsf{0.839_{0.001}}$	&$\mathsf{0.876_{0.002}}$	&$\mathsf{0.890_{0.002}}$\\
	        &OE$\downarrow$	&$\mathsf{0.359_{0.007}}$	&$\mathsf{0.296_{0.003}}$	&$\mathsf{0.249_{0.006}}$	&$\mathsf{0.337_{0.005}}$	&$\mathsf{0.286_{0.004}}$	&$\mathsf{0.482_{0.040}}$	&$\mathsf{0.410_{0.003}}$	&$\mathsf{0.272_{0.008}}$	&$\mathsf{0.253_{0.007}}$\\
	        &Cov$\downarrow$	&$\mathsf{0.358_{0.005}}$	&$\mathsf{0.334_{0.009}}$	&$\mathsf{0.303_{0.002}}$	&$\mathsf{0.339_{0.005}}$	&$\mathsf{0.309_{0.002}}$	&$\mathsf{0.421_{0.021}}$	&$\mathsf{0.364_{0.002}}$	&$\mathsf{0.298_{0.003}}$	&$\mathsf{0.278_{0.004}}$\\
	        \rowcolor{LGray}&Ave.R$\downarrow$	&7.17 	&5.17 	&2.33 	&5.83 	&4.17 	&9.00 	&7.67 	&2.50 	&1.17 \\
\bottomrule
\end{tabular}
\vspace{-0.2cm}
\end{table*}

\begin{table*}[t]

\renewcommand{\arraystretch}{1.1}
\caption{Ablation results of our RANK on three datasets with 50\% missing views, 50\% missing labels. The Backbone denotes our RANK without $\mathcal{L}_{re}$, $\mathcal{L}_{ma}$, and $\mathcal{L}_{ge}$. The dark color represents our full version of RANK.}
\label{table:ablation}%
	\begin{tabularx}{\textwidth}{cccc|cccc|cccc|cccm{0.5cm} }
		\toprule
		\multicolumn{4}{c|}{Method}&\multicolumn{4}{c|}{Corel5k}&\multicolumn{4}{c|}{ESPGame}&\multicolumn{4}{c}{IAPRTC12}\\
         Backbone &$\mathcal{L}_{re}$ &$\mathcal{L}_{ma}$ &$\mathcal{L}_{ge}$  &AP$\uparrow$&1-HL$\uparrow$&1-RL$\uparrow$&AUC$\uparrow$&AP$\uparrow$&1-HL$\uparrow$&1-RL$\uparrow$&AUC$\uparrow$&AP$\uparrow$&1-HL$\uparrow$&1-RL$\uparrow$&AUC$\uparrow$\\
         \midrule
         \underline{\Checkmark}&~&~&~&0.381  & 0.987  & 0.897  & 0.900   & 0.286  & 0.983  & 0.828  & 0.833   & 0.309  & 0.981  & 0.864  & 0.866   \\
         \underline{\Checkmark}&\underline{\Checkmark}&~&~&0.387  & 0.987  & 0.900  & 0.903   & 0.292  & 0.983  & 0.831  & 0.835   & 0.314  & 0.981  & 0.869  & 0.871   \\ 
         \underline{\Checkmark}&~&\underline{\Checkmark}&~&0.368  & 0.987  & 0.888  & 0.891   & 0.278  & 0.982  & 0.819  & 0.824   & 0.299  & 0.981  & 0.857  & 0.859   \\
         \underline{\Checkmark}&~&~&\underline{\Checkmark}&0.389  & 0.987  & 0.899  & 0.902   & 0.285  & 0.983  & 0.829  & 0.834   & 0.313  & 0.981  & 0.869  & 0.871   \\
         \underline{\Checkmark}&\underline{\Checkmark}&\underline{\Checkmark}&~&0.391  & 0.987  & 0.898  & 0.901   &0.295	&0.983	&0.835	&0.840   & 0.323  & 0.981  & 0.874  & 0.877   \\
         \underline{\Checkmark}&~&\underline{\Checkmark}&\underline{\Checkmark}&0.410  & 0.988  & 0.908  & 0.911   & 0.301  & 0.983  & 0.841  & 0.845   & 0.334  & 0.981  & 0.883  & 0.883   \\
         \underline{\Checkmark}&\underline{\Checkmark}&~&\underline{\Checkmark}&0.403  & 0.987  & 0.905  & 0.907   & 0.291  & 0.983  & 0.832  & 0.837   & 0.321  & 0.981  & 0.874  & 0.876   \\
         \rowcolor{Gray}\underline{\Checkmark}&\underline{\Checkmark}& \underline{\Checkmark}& \underline{\Checkmark}&0.427	&0.988	&0.913	&0.916			&0.314	&0.983	&0.848	&0.853			&0.346	&0.981	&0.888	&0.889  \\
         \midrule
		 \multicolumn{4}{c|}{RANK with equal weight}& 0.405  & 0.987  & 0.907  & 0.909 & 0.306  & 0.983  & 0.842  & 0.845  & 0.334  & 0.981  & 0.881  & 0.882  \\
		 \multicolumn{4}{c|}{RANK with fixed weight}& 0.407  & 0.987  & 0.907  & 0.910 & 0.307  & 0.983  & 0.842  & 0.846  & 0.334  & 0.981  & 0.883  & 0.883  \\
		 \multicolumn{4}{c|}{RANK with unsup-cont}&0.413	&0.988	&0.910	&0.912			&0.310	&0.983	&0.844	&0.848			&0.337	&0.981	&0.882	&0.883\\
         \multicolumn{4}{c|}{RANK w/o $\mathcal{L}_{mcce}$}&0.409 	&0.988 	&0.896 	&0.899 &0.307 	&0.983 	&0.842 	&0.847 &0.336 	&0.981 	&0.880 	&0.881   \\
         \multicolumn{4}{c|}{RANK w/o $\mathcal{L}_{mcce}$ and $\mathcal{D}$}&0.400  & 0.988  & 0.889  & 0.891  & 0.306  & 0.983  & 0.837  & 0.841  & 0.329  & 0.981  & 0.874  & 0.876  \\ 
         \bottomrule
     \end{tabularx}
     \vspace{-0.2cm}
\end{table*}

\begin{figure}[t]
\centering
\subfloat[Corel5k]{
\includegraphics[width=0.24\textwidth]{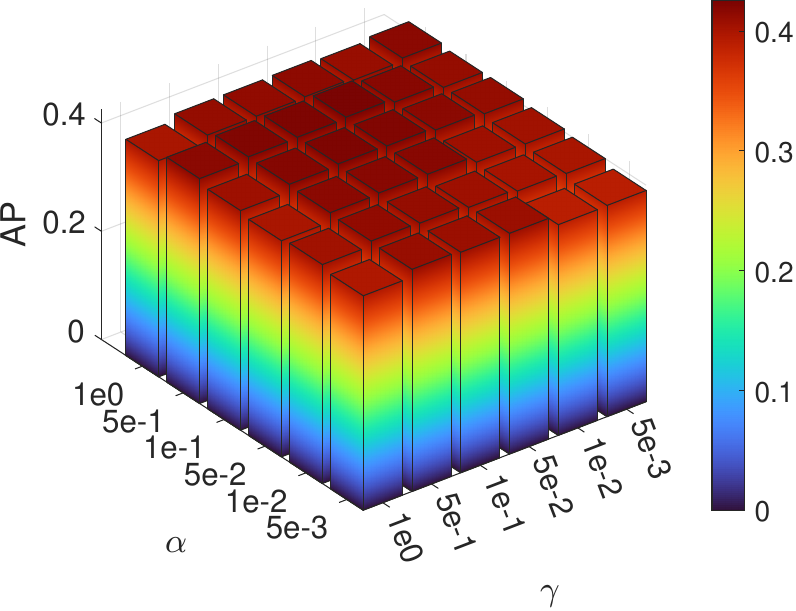}
}
\subfloat[Pascal07]{
\includegraphics[width=0.24\textwidth]{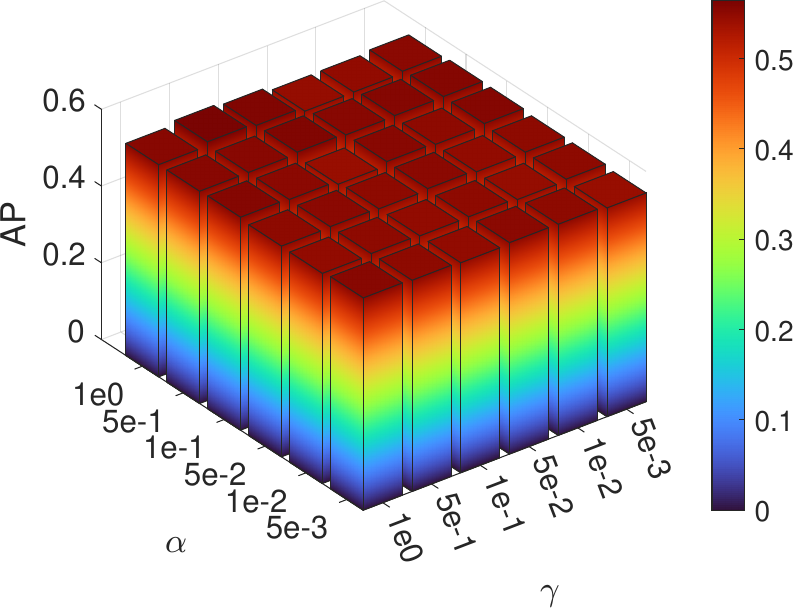}
}
\quad
\subfloat[Corel5k]{
\includegraphics[width=0.24\textwidth]{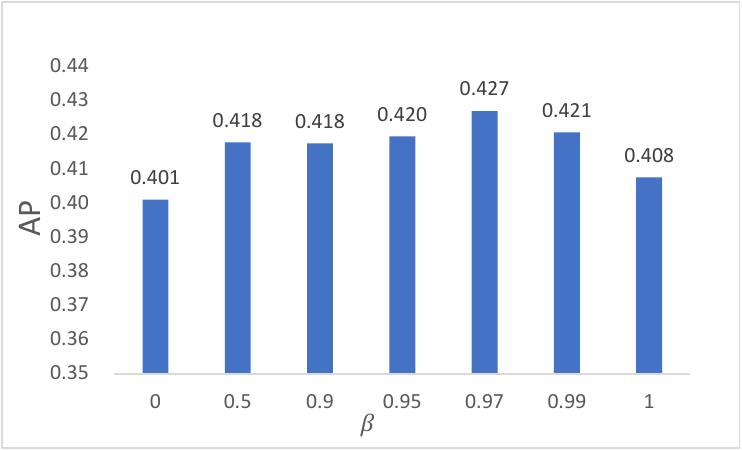}
}
\subfloat[Pascal07]{
\includegraphics[width=0.24\textwidth]{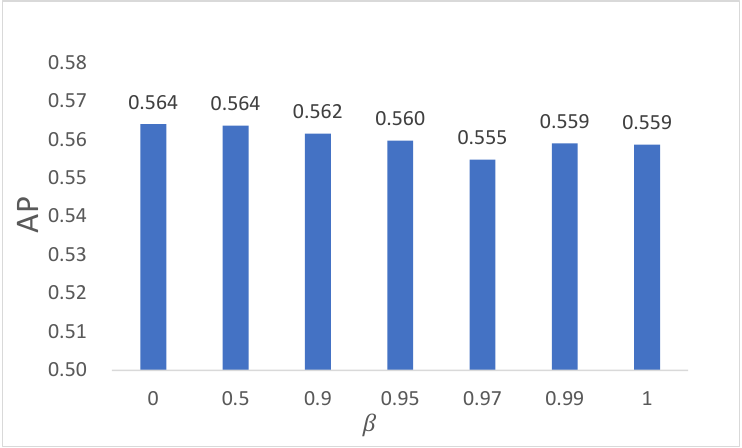}
}
\caption{AP v.s. $\alpha$ and $\gamma$ on the (a) Corel5k and (b) Pascal07 datasets; AP v.s. $\beta$ on the (c) Corel5k and (d) Pascal07 datasets. All datasets are with 50\% available labels, 50\% missing instances, and 70\% training samples.}
\label{fig:para}
\end{figure}

\subsection{Experimental Results}
In Table \ref{table:inc}, we report the experimental results of nine methods on the five datasets with 50\% missing instances, 50\% missing labels, and 70\% training samples. The bottom right decimal is the standard deviation. Besides, we also perform experiments on the datasets with 30\% view and label missing rates (Fig. \ref{fig:alla}), with full views and 50\% label missing rate (Fig. \ref{fig:allb}), and with full labels and 50\% view missing rate (Fig. \ref{fig:allc}). In order to study the impact of different view or label missing rates on our model, we fix one missing rate at 50\%, adjust the other missing rate to [0 30\% 50\% 70\%] respectively, and record the results in Fig. \ref{fig:rates}. From Table \ref{table:inc}, Fig. \ref{fig:rates}, and Fig. \ref{fig:alla}, it is easy to obtain the following observations:
\begin{itemize}
\item From Table \ref{table:inc}, the performance of our RANK is impressive, achieving \textit{first-place} results on all metrics across all datasets, when the dataset has a large proportion of missing data. For example, on the Pascal07 dataset, our method leads the second-best CDMM by about 5 percentage points in AP; on the ESPGame dataset, our RANK outperforms the second-best DICNet by 2 percentage points in OE.
\item As shown in Fig. \ref{fig:rates}, both missing views and labels will weaken the performance of our model, and when the missing rate increases, the performance will gradually decrease. In addition, at the same missing rate, incomplete views have a greater negative impact on the model than missing labels.
\item From Fig. \ref{fig:rates}, we can find an interesting phenomenon: When the view-missing rate exceeds 50\%, the model's performance will drop significantly. One possible reason is that the number of effective features that can be used for multi-view aggregation and intra-view structure preservation is too small, which weakens the advantage of our label-driven multi-view contrastive learning strategy. On the other hand, more than 50\% missing labels do not cause a huge loss in performance, which is attributed to the sparsity of these datasets' label matrices .
\item It can be intuitively seen from Fig. \ref{fig:all} that under different missing rates, our RANK can still achieve the best performance compared to other advanced methods. Besides, some methods designed for complete views or labels have huge performance gains in the case of full views or labels, such as CDMM and LVSL.
\end{itemize}

Although our RANK is designed for iM3C, it can also handle the complete multi-view multi-label dataset by setting all elements of the two missing indicator matrices to $1$. Table \ref{table:com} lists the nine methods' results under the complete multi-view multi-label datasets, from which we can see that our method still achieves the best ranking in Fig. \ref{fig:para}. The experimental results confirm the good robustness of our method in coping with multi-view multi-label data with different missing rates.
\begin{table*}[h!]
\center

	\centering
	\caption{Running time of training and testing phases of the nine methods on the Corel5k dataset with 70\% training samples. (Unit: second)}
	\label{table:effi}
	\resizebox{0.9\textwidth}{!}{
		\begin{tabular}{cccccccccc}
			\toprule
			\diagbox{Phase}{Method}     & C2AE & GLOCAL&CDMM  & DM2L &LVSL&iMVWL &NAIM3L&DICNet&RANK    \\ \midrule
			Training     &    170.24    &    154.42   &16.02 	&    713.37 &63.73	&165.82& 143.63 &313.89 &450.51 \\
			Test        &    0.04    &    0.89   &1.73  	&  0.04 &0.64	&0.02	&0.01 &0.05   &0.35\\\bottomrule
	\end{tabular}}
	
\end{table*}

\subsection{Ablation Experiments}
\label{sec:abe}
In our RANK, four modules and their corresponding loss functions are reviewed: multi-view autoencoder framework ($\mathcal{L}_{re}$), label-driven multi-view contrastive learning ($\mathcal{L}_{ma}$ and $\mathcal{L}_{ge}$), quality discriminator ($\mathcal{L}_{qd}$), and multi-label collaborative classification ($\mathcal{L}_{mcce}$). In order to further study the effectiveness of each component, in this section we conduct ablation experiments on RANK from four aspects: 
\begin{enumerate}
 \item We remove the three detachable losses, i.e., $\mathcal{L}_{re}$, $\mathcal{L}_{ma}$, and $\mathcal{L}_{ge}$ to obtain the \textit{backbone} model. And combine them separately to study the gain effects of these losses on the \textit{backbone} network; 
 \item To study the contribution of quality discriminator to multi-view weighted fusion, we replace our quality discriminator $\mathcal{D}$ and dynamically weighted fusion module by the traditional fusion approach like Eq. (\ref{eq:wfusion}): $\bar{z_i}=\sum_{v=1}^{m}a_vz_{i}^{(v)}$, which is commonly used in deep multi-view learning \cite{trosten2021reconsidering,liu2023dicnet,wen2020dimc}. In our experiments, two variants for incomplete multi-view data are adopted: (1) Treating all views as equally important: $ \bar{\mathbf{Z}}_{i,:} = \big(\sum\limits_{v=1}^{m}\mathbf{Z}_{i,:}^{(v)}\mathbf{W}_{i,v}\big)\big/\sum\limits_{v=1}^{m}\mathbf{W}_{i,v}$, where $\mathbf{W}$ is added to mask unavailable instances; (2) Introducing learnable view-level weights $\{a_v\}_{v=1}^{m}$  that sum to 1 and assigning the same weight to all instances in the same view: $ \bar{\mathbf{Z}}_{i,:} = \big(\sum\limits_{v=1}^{m}a_v\mathbf{Z}_{i,:}^{(v)}\mathbf{W}_{i,v}\big)\big/\sum\limits_{v=1}^{m}\mathbf{W}_{i,v}$. We call the two degraded versions ``RANK with equal weight'' and ``RANK with fixed weight'', respectively;
 \item In Section \ref{sec:l_d_cont}, we propose to replace the classical contrastive learning paradigm by a multi-view aggregation item and a label-driven graph embedding learning item, i.e., $\mathcal{L}_{ma}$ and $\mathcal{L}_{ge}$. To assess the effectiveness of the proposed supervised multi-view contrastive learning, we restore it to the classic unsupervised contrastive learning like Eq. (\ref{eq:cl2}) used in existing methods \cite{liu2023dicnet,xu2022multi}, and the new version is named as ``RANK with unsup-cont'';

 \item For the proposed $\mathcal{L}_{mcce}$, we also use its common version ``RANK with $\mathcal{L}_{mbce}$'' like Eq. (\ref{eq:mbce}) (In fact, a modified version of it is used for incomplete labels, see \cite{liu2023dicnet}), which is also wildly used in multi-label classification \cite{hang2021collaborative,lyu2022beyond}.
\end{enumerate}

In Table \ref{table:ablation}, we list the results of the ablation experiments w.r.t. the above two aspects. For the first part of the results, the full version of RANK achieves the best performance. In general, the absence of different losses degrades the performance to a certain extent, except for the combination of ‘Backbone+$\mathcal{L}_{ma}$’. This involves two deep-seated questions with respective to view alignment operation commonly used in deep multi-view learning \cite{hang2021collaborative}: 1) Is aligning multiple views necessarily beneficial? Clearly, our ablation experiments negate this. Blindly forcing multiple views to be close to each other has the potential to induce better views to learn towards the poorer views, possibly weakening the overall performance of the model. Therefore, the combination of multi-view aggregation loss with reconstruction loss and graph embedding learning loss can give full play to the advantages of multi-view consistency aggregation strategy. 2) Is it beneficial to align multiple views at any time? The answer is still no. Concretely, we expect to align multiple views under these premises: the view-specific features extracted by the encoders maintain the key properties of the views; the geometric structure inside the view tends to be stable. This is why we design an incremental coefficient for $\mathcal{L}_{ma}$. The results in the second part illustrate that both our quality discriminator and multi-label collaborative cross-entropy loss have positive contributions to the model, especially reflecting the importance of fine-grained quality evaluation for multi-view dynamic fusion.

\subsection{Parameters Study}
\label{sec:ps}
In our RANK, the setting of parameters $\alpha$, $\beta$, and $\gamma$ has a direct impact on the model. In this subsection, we perform experiments on two datasets, Corel5k and Pascal07 with 50\% missing instances and 50\% missing labels, to study the sensitivity of our model to the three parameters. As shown in Fig. \ref{fig:para}, our RANK is relatively insensitive to the selection of parameters $\alpha$ and $\gamma$. For example, selecting from a large parameter range, i.e., $\alpha \in [5e-1, 1e-2], \gamma \in [5e-1, 5e-2]$ for Corel5k dataset and $\alpha \in [5e-3, 1], \gamma \in [5e-1, 1e-1]$ for Pascal07 dataset, our method can easily achieve a good result. As for the parameter $\beta$, it controls the growth rate of the multi-view aggregation loss $\mathcal{L}_{ma}$ with the training process. From Fig. \ref{fig:para}, we can intuitively see that for the Corel5k dataset, the best performance can be obtained when a larger $\beta$ is selected, while the opposite is true for the Pascal07 dataset. For this phenomenon, combined with the analysis in Section \ref{sec:abe}, we believe that due to the small number of categories, the model needs fewer iterations (no more than 20 epochs) to fit the training set on the Pascal07 dataset, so setting a lower $\beta$ is beneficial to play the role of multi-view aggregation loss. On the contrary, for the Corel5k dataset with a large number of categories (required epochs are about 200), a larger $\beta$ can help the model further align all views in the later stage of training. In general, we recommend setting $\beta$ to $0$ for Pascal07 and Mirflickr datasets, and $\beta$ to $0.97$ for other three datasets.

\subsection{Time Cost Analysis}

To evaluate the running time of our RANK during the training and testing phases, we record the time cost of these nine methods on the Corel5k dataset with 70\% training samples in Table \ref{table:effi}. To avoid the bias caused by the various condition settings, we keep their convergence conditions unchanged to get as objective results as possible. For methods that work with a single view at a time, we record the total training time for all views and the testing time for a single view. For DICNet and RANK, we record the time cost of 100 training epochs. Besides, to keep hardware resources consistent, all experiments are conducted on the same computing platform. From results in Table \ref{table:effi}, we can observe that although our RANK spends more time in the training phase, it achieved better performance than other methods.

\section{Conclusion}
In this paper, we first provide an in-depth analysis of the drawbacks of existing contrastive multi-view learning methods, namely the category conflicts brought about by separating false negative pairs. And a label-driven multi-view contrastive learning strategy is proposed from both inter-view and intra-view levels to solve this problem, including multi-view aggregation and graph embedding learning. Second, in view of the fact that existing methods do not consider view balance or only consider view-level weight distribution, we innovatively propose a quality-aware sub-network to evaluate the view quality of each sample in a fine-grained and customized manner. Finally, we extend the widely used multi-label cross-entropy loss with the label correlation information to improve the discriminative power of classifier. Experiments performed on five popular multi-view multi-label datasets strongly demonstrate that our RANK outperforms the existing state-of-the-art methods in both missing views and labels as well as complete data. And ablation experiments strongly confirm the validity of each component of our RANK.


%



\ifCLASSOPTIONcompsoc
\fi


\ifCLASSOPTIONcaptionsoff
  \newpage
\fi



\bibliographystyle{IEEEtran}
\bibliography{IEEEabrv}
%
%
%

%
\vspace{-1cm}
\begin{IEEEbiography}[{\includegraphics[width=1in,clip,keepaspectratio]{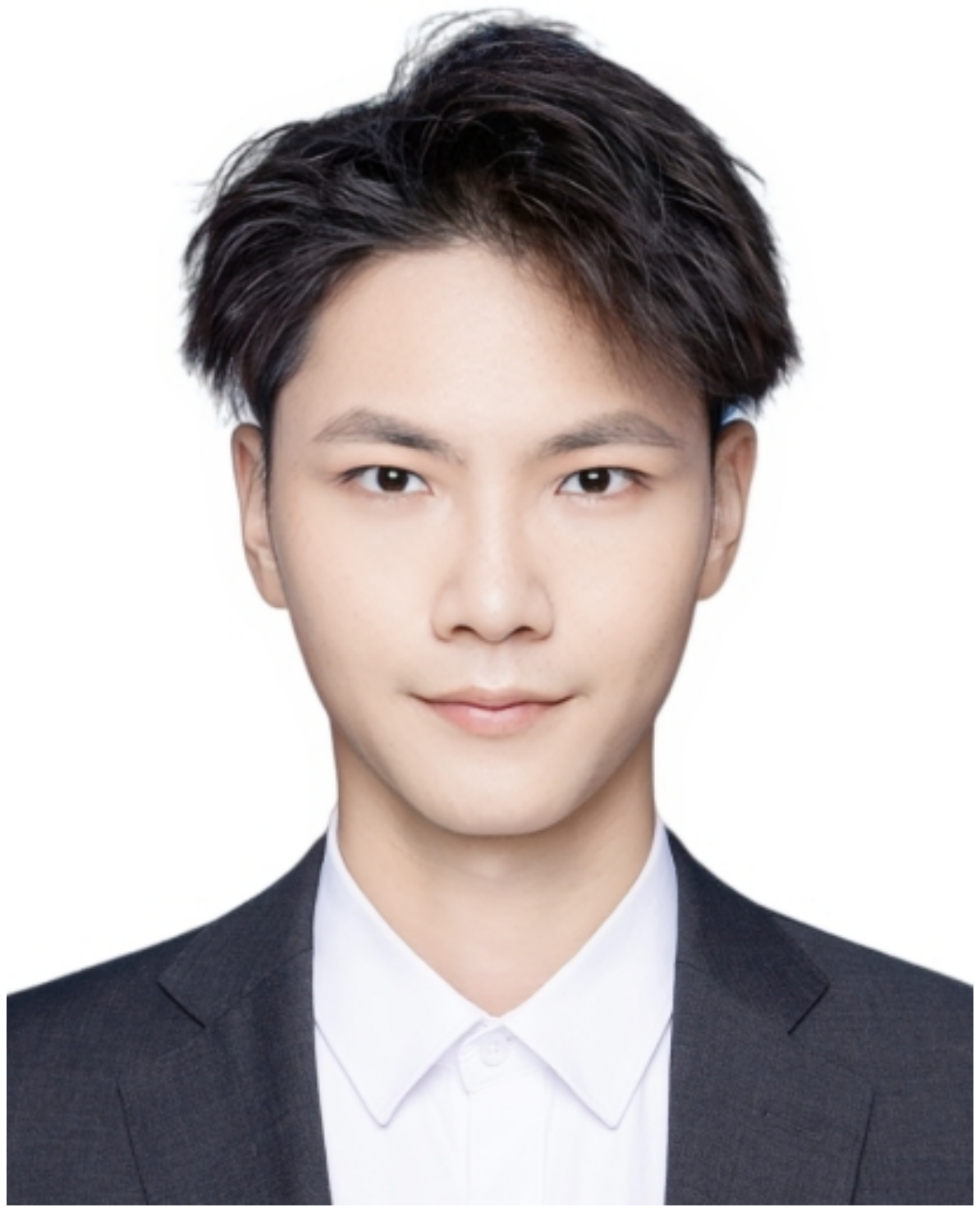}}]{Chengliang Liu}
(Member, IEEE) received the PhD degree in computer science from Harbin Institute of Technology, Shenzhen, China, in 2024, the M.S. degree in computer science from the Huazhong University of Science and Technology, Wuhan, China, in 2020, and the B.S. degree in computer science from the Jilin University, Changchun, China, in 2018. His research interests include machine learning and computer vision, especially multi-modal learning. Dr. Liu was honored with the Distinguished Paper Award at AAAI 2023, and he received the Harbin Institute of Technology’s Outstanding Dissertation Award. He served as the reviewer/ PC member for several top-tier journals and conferences, such as TIP, TCYB, TNNLS, CVPR, ICCV, ICML, NeurIPs, ICLR, AAAI, and IJCAI. His homepage: \url{https://justsmart.github.io}.
\end{IEEEbiography}
\begin{IEEEbiography}[{\includegraphics[width=1.0in,clip,keepaspectratio]{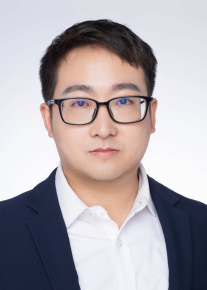}}]{Jie Wen}
 (Senior Member, IEEE)  received the Ph.D. degree in Computer Science and Technology at Harbin Institute of Technology, Shenzhen in 2019. He is currently an Associate Professor at the School of Computer Science and Technology, Harbin Institute of Technology, Shenzhen. His research interests include image and video enhancement, pattern recognition, and machine learning. He has authored or co-authored more than 100 technical papers at prestigious international journals and conferences, including the TNNLS, TIP, TCYB, NeurIPS, ICML, CVPR, AAAI, IJCAI, ACM MM, etc. He serves as an \textbf{Associate Editor} of \textit{IEEE Transactions on Image Processing}, \textit{IEEE Transactions on Information Forensics and Security}, \textit{Pattern Recognition}, and \textit{International Journal of Image and Graphics}, an \textbf{Area Editor} of \textit{Information Fusion}. He also served as the \textbf{Area Chair} of \textit{ACM MM} and \textit{ICML}. He was selected for the `World's Top 2\% Scientists List' in 2021-2024. One paper received the `distinguished paper award’ from AAAI’23. For more information, please refer to the homepage: \url{https://sites.google.com/view/jerry-wen-hit/home}.
\end{IEEEbiography}
\begin{IEEEbiography}[{\includegraphics[width=1.0in,clip,keepaspectratio]{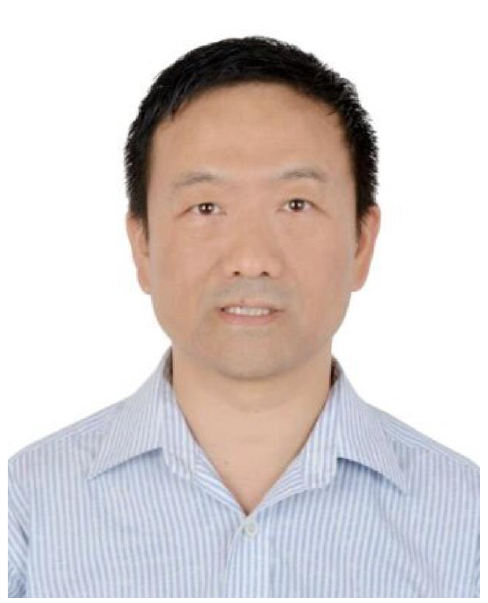}}]{Yong Xu}
(Senior Member, IEEE) received his B.S. degree, M.S. degree in 1994 and 1997, respectively. He received the Ph.D. degree in Pattern Recognition and Intelligence system at NUST (China) in 2005. He is currently a Professor with the School of Computer Science and Technology, Harbin Institute of Technology (HIT), Shenzhen. His research interests include pattern recognition, deep learning, biometrics, machine learning and video analysis. He has published over 70 papers in top-tier academic journals and conferences. His articles have been cited more than 5,800 times on the Web of Science, and 15,000 times in the Google Scholar. He has served as a Co-Editors-in-Chief of the International Journal of Image and Graphics, an Associate Editor of the \textit{CAAI Transactions on Intelligence Technology} and \textit{Neural Network}. 
\end{IEEEbiography}
\begin{IEEEbiography}[{\includegraphics[width=1.0in,clip,keepaspectratio]{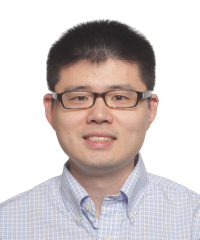}}]{Bob Zhang}
(Senior Member, IEEE) received the B.A. degree in computer science from York University, Toronto, ON, Canada, in 2006, the M.A.Sc. degree in information systems security from Concordia University, Montreal, QC, Canada, in 2007, and the Ph.D. degree in electrical and computer engineering from the University of Waterloo, Waterloo, ON, Canada, in 2011. After graduating from the University of Waterloo, he remained with the Center for Pattern Recognition and Machine Intelligence. He was a Post-Doctoral Researcher with the Department of Electrical and Computer Engineering, Carnegie Mellon University, Pittsburgh, PA, USA. He is currently an Associate Professor with the Department of Computer and Information Science, University of Macau, Taipa, Macau. His research interests include biometrics, pattern recognition, and image processing. He is a Technical Committee Member of the IEEE Systems, Man, and Cybernetics Society and an Associate Editor of IET Computer Vision.
\end{IEEEbiography}
\vskip -2cm
\begin{IEEEbiography}[{\includegraphics[width=1.0in,clip,keepaspectratio]{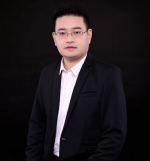}}]{Liqiang Nie}
(Senior Member, IEEE) IAPR Fellow, is currently the dean with the School of Computer Science and Technology, Harbin Institute of Technology (Shenzhen). He received his B.Eng. and Ph.D. degree from Xi’an Jiaotong University and National University of Singapore (NUS), respectively. His research interests lie primarily in multimedia content analysis and information retrieval. Dr. Nie has co-/authored more than 100 CCF-A papers and 5 books, with 15k plus Google Scholar citations. He is an AE of IEEE TKDE, IEEE TMM, IEEE TCSVT, ACM ToMM, and Information Science. Meanwhile, he is the regular area chair or SPC of ACM MM, NeurIPS, IJCAI and AAAI. He is a member of ICME steering committee. He has received many awards over the past three years, like ACM MM and SIGIR best paper honorable mention in 2019, the AI 2000 most influential scholars2020, SIGMM rising star in 2020, MIT TR35 China 2020, DAMO Academy Young Fellow in 2020, SIGIR best student paper in 2021, first price of the provincial science and technology progress award in 2021(rank 1), and provincial youth science and technology award in 2022. Some of his research outputs have been integrated into the products of Alibaba, Kwai, and other listed companies.
\end{IEEEbiography}
\begin{IEEEbiography}[{\includegraphics[width=1.0in,clip,keepaspectratio]{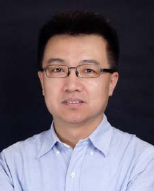}}]{Min Zhang}
(Member, IEEE) received the bachelor’s and Ph.D. degrees in computer science from Harbin Institute of Technology, Harbin, in 1991 and 1997, respectively. He currently serves as the Assistant to the President at Harbin Institute of Technology, Shenzhen, China. His current research interests include natural language processing, multi-model, artificial intelligence, and cryptology.
\end{IEEEbiography}








\end{document}